\documentclass[10pt,journal]{IEEEtran}

\usepackage{mathrsfs}

\usepackage{latexsym}
\usepackage{amssymb}
\usepackage{graphicx}
\usepackage{amsmath}
\usepackage{epsfig}
\usepackage{latexsym}
\usepackage{indentfirst}
\usepackage{amsthm}
\usepackage{subfigure}
\usepackage{amsfonts}
\usepackage{picins}

\ifCLASSOPTIONcompsoc
\else
\fi
\ifCLASSINFOpdf
\else
\fi

\usepackage{graphicx}

\usepackage{multirow}
\newtheorem{theorem}{Theorem}
\newtheorem{assumption}{Assumption}
\newtheorem{corollary}{Corollary}
\newtheorem{lemma}{Lemma}

\newtheorem{proposition}{Proposition}

\graphicspath{{figures/}}

\begin{document}
%
\title{Generalization and Expressivity for Deep Nets}
%
%
%
%

\author{Shao-Bo Lin
\IEEEcompsocitemizethanks{\IEEEcompsocthanksitem S. Lin is with the
Department of Mathematics, Wenzhou University, Wenzhou, China and
also with the Department of Mathematics, City University of
Hong Kong, Kowloon, Hong Kong, China.}}

\maketitle \IEEEcompsoctitleabstractindextext{
\begin{abstract}
Along with the rapid development of deep learning in practice,
theoretical explanations for its  success become  urgent.
Generalization and expressivity  are two widely used measurements to
quantify theoretical behaviors of deep learning. The expressivity
focuses on finding functions expressible by deep nets but cannot be
approximated by shallow nets with the similar number of neurons. It
usually implies the large capacity. The generalization aims at
deriving fast learning rate for deep nets. It usually requires small
capacity to reduce the variance.
 Different from previous
studies  on deep learning,  pursuing  either   expressivity or
generalization,    we take both factors into account to explore
theoretical advantages of deep nets. For this purpose,  we construct
a deep net with two hidden layers possessing excellent expressivity
in terms of localized and sparse approximation. Then, utilizing the
well known covering number to measure the capacity, we find that
deep nets possess excellent expressive  power (measured by localized
  and sparse approximation) without essentially enlarging the
capacity of shallow nets. As a consequence, we derive  near optimal
learning rates for implementing empirical risk minimization (ERM) on
the constructed deep nets. These results theoretically exhibit the
  advantage of deep nets  from    learning theory viewpoints.
\end{abstract}

}

\maketitle

\IEEEdisplaynotcompsoctitleabstractindextext

\IEEEpeerreviewmaketitle


\section{Introduction}\label{Sec.Introduction}
Technological innovations on data mining bring  massive data   in
diverse areas of modern scientific research \cite{Zhou2014}.   Deep
learning  \cite{Hinton2006,Bengio2009} is recognized to be a
state-of-the-art scheme to take advantage of massive data, due to
their {\it unreasonable effective} empirical evidence. Theoretical
verifications   for such effectiveness of deep learning is a hot
topic in recent years' statistical and machine learning
\cite{Goodfellow2016}.

One of the most important reasons for the success of deep learning
is the utilization of deep nets, a.k.a.,   neural networks with more
than one  hidden layers. In the classical neural network
approximation literature \cite{Pinkus1999}, deep nets were shown to
outperform shallow nets, i.e., neural networks with one hidden
layer, in terms of providing localized approximation  and breaking
through some lower bounds for shallow nets approximation. Besides
these classical assertions, recent focus
\cite{Kurkova2013,Eldan2015,Telgarsky2016,Mhaskar2016,Lin2017} on
deep nets approximation is to provide various functions expressible
for deep nets but cannot be approximated by shallow nets with
similar number of neurons. All these results present theoretical
verifications for the necessity of deep nets from  the approximation
theory viewpoint.

Since deep nets can approximate more functions than shallow nets,
the capacity of deep nets seems to be   larger than that of shallow
nets with similar number of neurons. This argument was recently
verified under some specified complexity measurements such as  the
number of linear regions \cite{Montufar2013}, Betti numbers
\cite{Bianchini2014}, number of monomials \cite{Delalleau2011} and
so on \cite{Raghu2016}. The large capacity of deep nets inevitably
comes with the downside of increased overfitting risk according to
the bias and variance trade-off principle \cite{Cucker2007}. For
example, deep nets {\it with finitely many neurons} were proved in
\cite{Maiorov1999b} to be capable of approximating arbitrary
continuous function within arbitrary accuracy, but the
pseudo-dimension \cite{Maiorov1999} for such  deep nets is infinite,
which usually leads to extremely large variance in the learning
process. Thus the existing necessity of deep nets in the
approximation theory community cannot be used directly to explain
the feasibility of deep nets  in
 machine learning.

In this paper, we aim at studying the learning performance for
implementing empirical risk minimization (ERM) on some specified
deep nets. Our analysis starts with the localized approximation
property as well as the sparse approximation ability of deep nets to
show their expressive power. We then conduct a refined estimate for
the covering number \cite{Zhou2003} of deep nets, which is closely
connected to learning theory \cite{Cucker2007}, to measure the
capacity. The result shows that, although deep nets possess
localized and sparse approximation while shallow nets fail, their
capacities measured by the covering number are similar, provided
there are comparable number of neurons in both nets. As a
consequence, we derive almost optimal learning rates for the
proposed ERM algorithms on deep nets when the so-called regression
function \cite{Cucker2007} is Liptchiz continuous. Furthermore,  we
prove that   deep nets can reflect the sparse property of the
regression functions via breaking through the established almost
optimal learning rates.
 All these results    show  that learning schemes based on deep
nets can learn more (complicated) functions than those based on
shallow nets.

 The rest of this paper is organized as
follows. In the next section, we present some results on the
expressivity and capacity of deep nets. These properties were
utilized in Section \ref{Sec.Main result} to show outperformance of
deep nets in the machine learning community. In Section
\ref{Sec.Related work}, we present some related work and
comparisons. In the last section, we draw a simple conclusion of our
work.

\section{Expressivity and Capacity}\label{Sec.Expressivity}

Expressivity \cite{Raghu2016} of deep nets usually means that deep
nets can represent some functions that cannot be approximated by
shallow nets with similar number of neurons. Generally speaking,
expressivity implies the large capacity of deep nets. In this
section, we  firstly show the expressivity of deep nets in terms of
localized  and sparse approximation,  and then prove that the
capacity measured by  covering number is not essentially enlarged
when the number of hidden layer increases.

\subsection{Localized approximation for deep nets}

Let
$$
       S_{\sigma,n}=\left\{\sum_{j=1}^nc_j\sigma(w_jx+\theta_j):c_j,\theta_j\in\mathbb
       R, w_j\in\mathbb R^d\right\}
$$
be the set of shallow nets with activation function $\sigma$ and $n$
neurons. Denote by $D_{\sigma_1,\sigma_2,n_1,n_2}$ the set of deep
nets with two hidden layers
$$
      g(x)=\sum_{k=1}^{n_2}c_k
      \sigma_2\left(\sum_{j=1}^{n_1}c_{k,j}\sigma_1(w_{k,j}x+\theta_{k,j})+\theta_k\right)
$$
where $c_k,c_{k,j},\theta_k,\theta_{k,j}\in\mathbb
       R, w_{k,j}\in\mathbb R^d$. The aim of this subsection is to show
the outperformance of $D_{\sigma_1,\sigma_2,n_1,n_2}$ over $
S_{\sigma,n}$ to verify the necessity of depth in providing
localized approximation.

The localized approximation of a neural network \cite{Chui1994}
shows that if the target function is modified only on a small subset
of the Euclidean space, then only a few neurons, rather than the
entire network, need to be retrained. As shown in Figure 1, a neural
network  with localized approximation should recognize the location
of the input in a small region.
\begin{figure}
\begin{center}
\includegraphics[height=5cm,width=7cm]{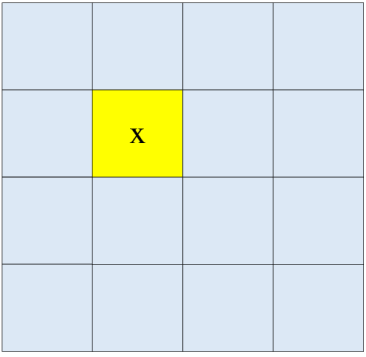}
 \caption{Localized approximation:realizing the location  of inputs
}
\end{center}
\label{Fig.Localization}
\end{figure}
Mathematically speaking, localized approximation means that for
arbitrary hypercube   $Q \subset\mathcal X$    and arbitrary
$n\in\mathbb N$, it is capable of finding a neural network $h$ such
that $
         \chi_Q=h,
$ where $\mathcal X$ is the input space and  $\chi_Q$ denotes the
indicator function of the set $Q$, i.e., $\chi_Q(x)=1$ when $x\in Q$
and $\chi_Q(x)=0$ when $x\notin Q$.

 Let  $d\geq 2$ and $\sigma_0$ be the heaviside function, i.e.
$ \sigma_0(t)=1$, when $t\geq 0$ and
 $\sigma_0(t)=0$ when $t<0$.
 It can be found in  \cite[Theorem 5]{Blum1991} (see also \cite{Chui1994,Pinkus1999})  that $S_{\sigma_0,n}$ cannot
provide localized approximation, implying that functions in
$S_{\sigma_0,n}$ with finite number of neurons cannot catch the
position information of the input. However, in the following, we
will construct a  deep net in $D_{\sigma_0,\sigma,2d,1}$ with some
activation function $\sigma$ and totally $2d+1$ neurons to recognize
the location of the input.

Let $\sigma:\mathbb R\rightarrow\mathbb R$ be
 a
sigmoidal function, i.e.,
$$
           \lim_{t\rightarrow +\infty }\sigma (t)=1,\quad \lim_{t\rightarrow -\infty
           }\sigma
           (t)=0.
$$
Then, for arbitrary $\varepsilon>0$,  there exists a
$K_\varepsilon:=K(\varepsilon,\sigma)>0$ depending only on $\sigma$
and $\varepsilon$ such that
\begin{eqnarray}\label{definition K for sigmoidal 1}
               \left\{ \begin{array}{cc} |\sigma(t)-1|<\varepsilon,&
               \mbox{if}\
                 t\geq K_\varepsilon,\\
                 |\sigma(t)|<\varepsilon,
                 &
                 \mbox{if}\
                 t\leq-K_\varepsilon.
                 \end{array}\right.
\end{eqnarray}
Let $\mathbb I^d:=[0,1]^d$. Denote by $\{A_{n, {\bf j}}\}_{{\bf
j}\in   \mathbb N_{n}^d}$ the cubic partition of $\mathbb I^d$  with
centers $\{\xi_{\bf j}\}_{{\bf j}\in   \mathbb N_{n}^d}$ and side
length $\frac1n$, where we write arbitrary vector ${\bf a}\in\mathbb
R^d$ as ${\bf a}=(a^{(1)},\dots,a^{(d)})^T$ and $\mathbb
N_{n}^d=\{1,2,\dots,n\}^d$ . Then, for $K>0$ and arbitrary ${\bf
j}\in\mathbb N_{n}^d$, we construct a deep net
$D_{\sigma_0,\sigma,2d,1}$ by
\begin{eqnarray}\label{NN for localization}
      &&N^*_{n,{\bf j},K}(x)
       :=
      \sigma\left\{2K\left[\sum_{\ell=1}^d\sigma_0\left[\frac1{2n}
      +x^{(\ell)}-\xi_{\bf j}^{(\ell)}\right]\right.\right.\nonumber\\
      &+&\left.\left.
       \sum_{\ell=1}^d\sigma_0\left[\frac1{2n}-x^{(\ell)}+\xi_{\bf j}^{(\ell)}\right]
      - 2d+\frac12 \right]\right\}.
\end{eqnarray}
In the following proposition proved in Appendix A, we show that deep
nets possess totally different property from shallow nets in
localized approximation.

\begin{proposition}\label{Proposition:localization}
For arbitrary $\varepsilon>0$, if  $N^*_{n,{\bf j},K_\varepsilon}$
is defined by (\ref{NN for localization}) with $K_\varepsilon$
satisfying (\ref{definition K for sigmoidal 1}) and $\sigma$ being a
non-decreasing sigmoidal function, then

\item{(a)} For arbitrary $x \notin A_{n,{\bf j}}$,
there holds $ N^*_{n,{\bf j},K_\varepsilon}(x) <\varepsilon.$

\item{(b)} For arbitrary $x\in   A_{n, {\bf j}},$ there holds
$1-N^*_{n,{\bf j},K_\varepsilon}(x)\leq \varepsilon$.
\end{proposition}

If we set $\varepsilon\rightarrow0$, Proposition
\ref{Proposition:localization} shows that $N^*_{n,{\bf
 j},K_\varepsilon}$ is an indicator function for $A_{n,{\bf j}}$, and
 consequently
provides localized approximation. Furthermore, as $n\rightarrow
\infty$, it follows from Proposition \ref{Proposition:localization}
that  $N^*_{n,{\bf
 j},K_\varepsilon}$ can recognize the location of $x$ in an
arbitrarily small region.
 In the prominent paper
\cite{Chui1994}, the  localized approximation property of deep nets
with two hidden layers and sigmoidal activation functions was
established in a weaker sense. The  difference between Proposition
\ref{Proposition:localization} and results in \cite{Chui1994} is
that we adopt the heaviside activation function in the first hidden
layer to guarantee the equivalence of $ N^*_{n,{\bf
j},K_\varepsilon}$ and $\chi_{_{A_{n,{\bf j}}}}$. In the second
hidden layer, it will be shown in Section II.C that some smoothness
assumptions should be imposed on the activation function to derive a
tight bound of the covering number. Thus, we do not recommend  the
use of heaviside activation. In short,   we require different
activation functions in different hidden layers to show excellent
expressivity and small capacity of deep nets.

Compared with shallow nets in $S_{\sigma_0,n}$, the constructed deep
net  $N^*_{n,{\bf j},K}$ introduces a   second hidden layer  to act
as a {\it judger}    to discriminate the location of inputs. Figure
2 below numerically exhibits the localized approximation of
$N^*_{n,{\bf j},K}$ with $n=4,d=2,K=10000$, $\xi_{\bf j}$ being the
center of the yellow zone in Figure 1 and $\sigma$ being the
logistic function, i.e., $\sigma(t)=\frac1{1+e^{-t}}$. As shown in
Figure 2, we can construct deep net that control a small region of
the input space but is independent of other regions. Thus, if the
target function changes only on a small region, then it is
sufficient to tune  a few neurons, rather than retrain  the entire
network. Since the locality of the data abound in  sparse coding
\cite{Olshausen1997}, statistical physics \cite{LinH2017} and image
processing \cite{Wang2004}, the localized approximation makes deep
nets be  effective and efficient in the related applications.
\begin{figure}
\begin{center}
\includegraphics[height=5cm,width=7cm]{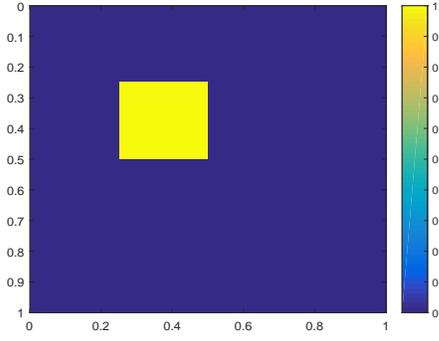}
 \caption{Localized approximation for the constructed deep net in (\ref{NN for
 localization})}
\end{center}
\label{Fig.NN L}
\end{figure}

\subsection{Sparse approximation for deep nets}
The localized approximation property of deep nets shows their power
to recognize functions defined on small regions. A direct
consequence is that deep nets can reflect the sparse property of the
target functions in the spacial domain.
 In this part, based on the
localized approximation property established in Proposition
\ref{Proposition:localization}, we focus on developing a deep net
with sparse approximation property in the spacial domain.

Sparseness in the spacial domain means that the response of some
actions happens  only on several small regions in the input space,
just as sparse coding \cite{Olshausen1997} purports to show. As
shown in Figure 3, sparseness studied in this paper means the
response (or function) vanishes in a large number of regions and
requires neural networks to recognize where the response does not
vanish.
\begin{figure}
\begin{center}
\includegraphics[height=5cm,width=7cm]{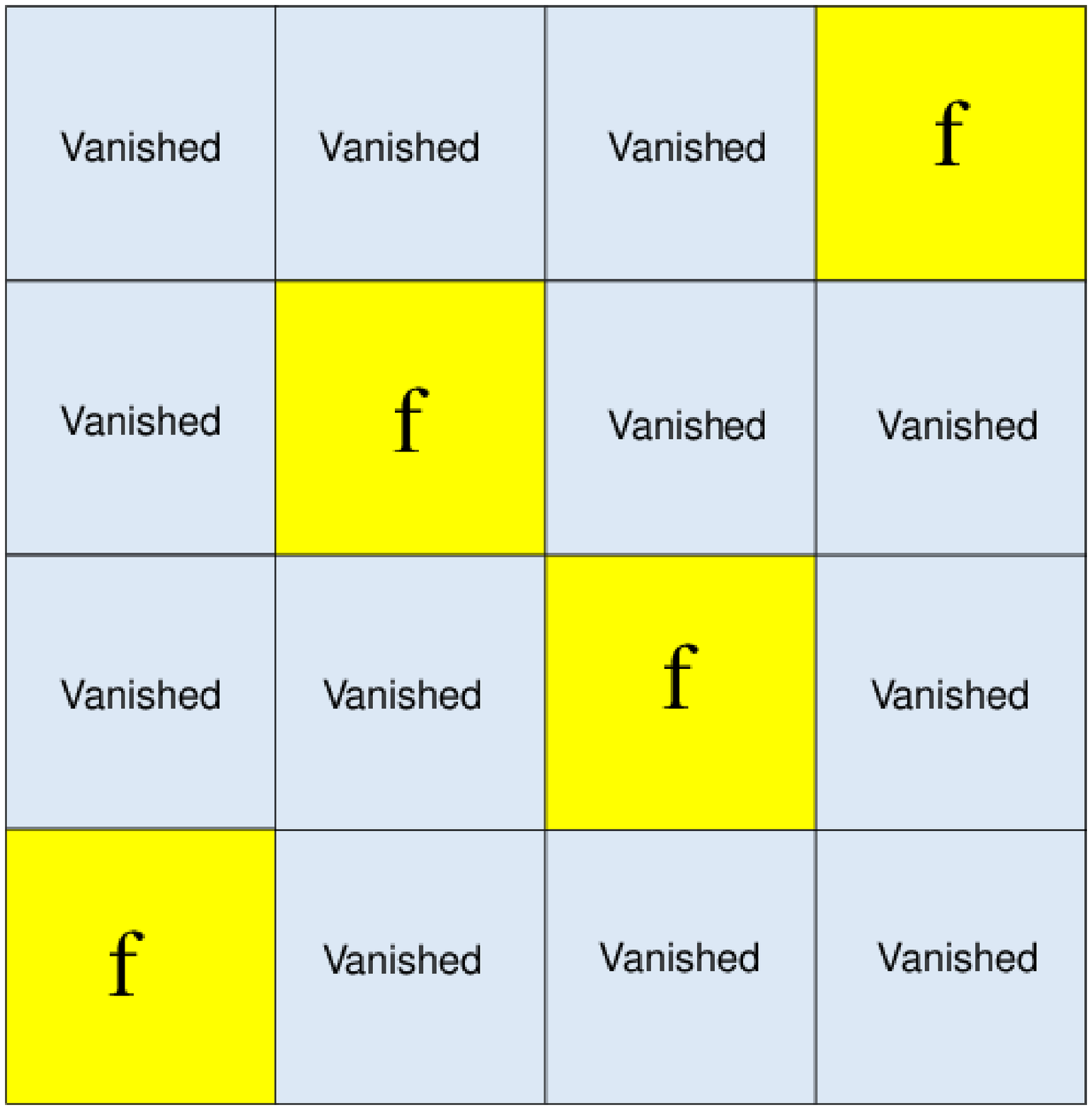}
 \caption{Sparseness in the spacial domain: an example of 4-sparse in 16 partitions function}
\end{center}
\end{figure}

Mathematically speaking, denote by  $\{B_{N, {\bf k}}\}_{{\bf k}\in
\mathbb N_{N}^d}$ the cubic partitions of $\mathbb I^d$ with center
$\zeta_{\bf k}$ and side length $\frac1N$.  For $s\in\mathbb N$ with
$s\leq N^d$, define
\begin{equation}\label{def lambda}
    \Lambda_s:=\left\{{\bf k}_\ell:
      {\bf k}_\ell\in \mathbb N_{N}^d, 1\leq \ell\leq s\right\}
\end{equation}
and
\begin{equation}\label{def support}
      S:=\cup_{{\bf k}\in \Lambda_s}B_{N,{\bf k}}.
\end{equation}
It is easy to see that $S$ contains  arbitrary regions consisting at
most $s$ sub-cubes (such as the yellow zones in Figure 3 with
$s=4$). We then say $S$ is a sparse subset of $\mathbb I^d$ of
sparseness $s$. For some function $f$ defined on $\mathbb I^d$, if
the support of $f$ is $S$, we then say that $f$ is $s$-sparse in
$N^d$ partitions.

As discussed above, the sparseness depends on the localized
approximation property. We thus can construct a deep net to embody
the spareness by the help of the constructed deep net in (\ref{NN
for localization}). For arbitrary $\varepsilon>0$   and
$\eta:=\{\eta_{\bf j}\}_{{\bf j}\in\mathbb N_n^d}$ with $ \eta_{{\bf
j}}\in A_{n,{\bf j}}$, define
\begin{equation}\label{Def.N2}
     N_{n,\eta,K_\varepsilon}(x):=\sum_{{\bf j}\in\mathbb N_n^d} f(\eta_{\bf j})N^*_{n,{\bf
     j},K_\varepsilon}(x),
\end{equation}
where $\{A_{n,{\bf j}}\}_{{\bf j}\in\mathbb N_n^d}$ is the cubic
partition   defined in the previous subsection. Obviously, we have
$N_{n,\eta,K_\varepsilon}\in D_{\sigma_0,\sigma,2d,n^d}$ which
possesses $n^d(2d+1)$ neurons.  In the following Proposition
\ref{Proposition:sparse approximation}, we will show that $
N_{n,\eta,K_\varepsilon}$ can embody the sparseness of the target
function by exhibiting a fast approximation rate which breaks
through the bottleneck of shallow nets.

For this purpose, we should  at first introduce some a-priori
information on the target function.   We say a function $f:\mathbb
I^d\rightarrow\mathbb R$ is $(r,c_0)$-Lipschitz if $f$ satisfies
\begin{equation}\label{lip}
          \left|f(x)-f(x')\right|\leq c_0\|x-x'\|^r,\qquad\forall\
          x,x'\in\mathbb I^d,
\end{equation}
where  $r,c_0>0$ and $\|x\|$ denotes the Euclidean norm of
  $x$.
Denote by $Lip^{(r,c_0)}$ the family of $(r,c_0)$-Lipschitz
functions satisfying (\ref{lip}). The Lipschitz property describes
the smoothness information of $f$ and has been adopted in vast
literature \cite{Chui1994,Maiorov1999,Pinkus1999,Lin2014b,Chui2016a}
to quantify the approximation ability of neural networks. Denote by
$Lip^{(N,s,r,c_0)}$ the set of all    $f\in Lip^{(r,c_0)}$ which is
$s$-sparse in $N^d$ partitions. It is easy to check that
$Lip^{(N,s,r,c_0)}$ quantifies both smoothness information and
sparseness in the spacial domain of the target function.

Then, we introduce the support set of $ N_{n,\eta,K_\varepsilon}$.
Note that the number of neurons of $ N_{n,\eta,K_\varepsilon}$
controls the side length of the cubic partition $\{A_{n, {\bf
j}}\}_{{\bf j}\in  \mathbb N_{n}^d}$, while $f$ is supported on $s$
cubes in  $\{B_{N, {\bf k}}\}_{{\bf k}\in \mathbb N_{N}^d}$. Since
$\{B_{N, {\bf k}}\}_{{\bf k}\in \mathbb N_{N}^d}$ is fixed, we need
to tune $n$ such that the constructed deep net $
N_{n,\eta,K_\varepsilon}$ can recognize each $B_{N, {\bf k}}$ with
${\bf k}\in \mathbb N_{N}^d$. Under this circumstance, we take
$n\geq 4N$ and for each  ${\bf k}\in\mathbb N_{N}^d$, define
\begin{equation}\label{Def.over Lambda}
    \overline{\Lambda_{\bf k}}:=\{{\bf j}\in \mathbb N_n^d:A_{n,{\bf j}}\cap
    B_{N,{\bf k}}\neq\varnothing\}.
\end{equation}
The set $\bigcup_{{\bf k}\in\Lambda_s}\overline{\Lambda_{\bf k}}$
corresponds to the family of  cubes $A_{n,{\bf j}}$ where $f$ is not
vanished. Since each $A_{n,{\bf j}}$ can be recognized by $2d+1$
neuron of $N_{n,\eta,K_\varepsilon}$ as given in Proposition
\ref{Proposition:localization}, $\bigcup_{{\bf
k}\in\Lambda_s}\overline{\Lambda_{\bf k}}$ actually describes the
support of $N_{n,\eta,K_\varepsilon}$. With these helps, we exhibit
in the following proposition  that $N_{n,\eta,K_\varepsilon}$
possesses the spare approximation ability, whose proof will be
presented in Appendix A.

\begin{proposition}\label{Proposition:sparse approximation}
Let $\varepsilon>0$ and $N_{n,\eta,K_\varepsilon}$ be defined by
(\ref{Def.N2}). If  $f\in Lip^{(N,s,r,c_0)}$ with $N,s\in\mathbb N$,
$0<r\leq 1$ and $c_0>0$, $K_\varepsilon$ satisfies (\ref{definition
K for sigmoidal 1}), $\sigma$ is a non-decreasing sigmoidal function
and $\eta=\{\eta_{\bf j}\}_{{\bf j}\in\mathbb N_n^d}$ with $
\eta_{{\bf j}}\in A_{n,{\bf j}}$, then for arbitrary  $x\in\mathbb
I^d$, there holds
\begin{equation}\label{sparse approximation1}
    |f(x)-N_{n,\eta,K_\varepsilon}(x)|\leq
    2^{r/2}c_0n^{-r}+\|f\|_{L^\infty(\mathbb
    I^d)}n^d\varepsilon.
\end{equation}
Furthermore, if $n\geq 4N$, we have
\begin{equation}\label{sparse approximation2}
    | N_{n,\eta,K_\varepsilon}(x)| \leq
    \|f\|_{L^\infty(\mathbb
    I^d)}n^d\varepsilon,\qquad  \forall \ x\in\mathbb I^d\backslash
    \bigcup_{{\bf k}\in\Lambda_s}\overline{\Lambda_{\bf k}}.
\end{equation}
\end{proposition}

It can be derived from (\ref{sparse approximation1}) with $S=\mathbb
I^d$ and $\varepsilon\leq n^{-d-r}$ that the deep net constructed in
(\ref{Def.N2}) satisfies the well known Jackson-type inequality
\cite{Lin2014b} for multivariate functions. This property shows that
in approximating Lipschitz functions, deep nets perform at least not
worse than shallow nets \cite{Pinkus1999}. If additional sparseness
information is presented, i.e. $f\in Lip^{(N,s,r,c_0)}$ with
$s<N^d$,
  by setting $\varepsilon\rightarrow0$,
(\ref{sparse approximation2}) illustrates that for every
$x\in\mathbb I^d\backslash
    \bigcup_{{\bf k}\in\Lambda_s}\overline{\Lambda_{\bf k}}$,
$$
    N_{n,\eta,K_\varepsilon}(x)  \rightarrow
   0,
$$
implying the sparseness of $N_{n,\eta,K_\varepsilon}$ in the spacial
domain.
It should be highlighted that for each ${\bf k}\in \Lambda_s$  the
cardinality of $\overline{\Lambda_{\bf k}}$, denoted by
$|\overline{\Lambda_{\bf k}}|$, satisfies
\begin{equation}\label{Card over lam}
     |\overline{\Lambda_{\bf k}}|\leq
     \left(\frac{n}{N}+2\right)^d\leq\frac{2^dn^d}{N^d},\qquad\forall \
     n\geq 4N.
\end{equation}
Therefore, there are at least
$$
    (2d+1)n^d- (2d+1)\frac{ s2^dn^d}{N^d}= (2d+1)n^d\frac{N^d-2^ds}{N^d}
$$
neurons satisfying (\ref{sparse approximation2}), which is large
when $s$ is small with respect to $N^d$. The aforementioned sparse
approximation ability  reduces the complexity of deep nets in
approximating sparse functions, which   makes deep-net-based
learning breaks though some limitations
 of shallow-net-based learning, as shown in Section \ref{Sec.Main
 result}.

\subsection{Covering number of deep nets}
Proposition \ref{Proposition:localization} and Proposition
\ref{Proposition:sparse approximation} show  the expressive power of
deep nets. In this subsection, we exhibit that the capacity of deep
nets, measured by the well known covering number, is similar as that
of shallow nets, implying that deep nets can approximate more
functions than shallow nets but do not bring additional costs.

Let $B$ be a Banach space and $V$ be a compact set in $B$. Denote by
$\mathcal N(\varepsilon,V,B)$ the covering number \cite{Zhou2003} of
$V$ under the metric of $B$, which is the number of elements in
least $\varepsilon$-net of $V$. If $B=C(\mathbb I^d)$, the space of
continuous functions, we denote $\mathcal N(\varepsilon,V):=\mathcal
N(\varepsilon,V,C(\mathbb I^d))$ for brevity. The estimate of
covering number of shallow nets  is a classical research topic in
approximation and learning theory
\cite{Makovoz1996,Kurkova2007,Gyorfi2002,Maiorov2006a,Maiorov2006b}.
Our purpose is to present a refined estimate for the covering number
of deep nets to show whether there are additional costs required by
deep nets to embody the localized   and sparse approximation.

To this end, we focus on a special subset of
$D_{\sigma_1,\sigma_2,n_1,n_2}$ which consists the deep nets
satisfying Propositions \ref{Proposition:localization} and
\ref{Proposition:sparse approximation}. Let $g$ be a deep net with
two hidden layers defined by
\begin{eqnarray*}
     g(x)&=&\sum_{j=1}^{n^d}c_j\sigma\left(\sum_{\ell=1}^{d}\alpha_{j,\ell}\sigma_0
     \left(x^{(\ell)}+\beta_{j,\ell}\right)\right.\\
     &+&
     \left.\sum_{\ell=1}^{d}\alpha'_{j,\ell}\sigma_0
     \left(x^{(\ell)}+\gamma_{j,\ell}\right)+b_j\right),
\end{eqnarray*}
where
$c_j,b_j,\alpha_{j,\ell},\beta_{j,\ell},\gamma_{j,\ell}\in\mathbb
R$. Define $\Phi_{n,2d}$ be the family of such deep nets whose
parameters are bounded, i.e.,
\begin{eqnarray}\label{Hypothesis space}
     \Phi_{n,2d} &:=&  \big\{g:|c_j|\leq \mathcal C_n,
     |b_j|\leq\mathcal B_n, |\alpha_{j,\ell}|, \nonumber\\
     &&|\alpha'_{j,\ell}|\leq \Xi_n,
      \beta_{j,\ell},  \gamma_{j,\ell}\in\mathbb R\big\},
\end{eqnarray}
where $\mathcal B_n,\mathcal C_n$ and $\Xi_n,$  are positive
numbers. We can see  $N_{n,\eta,K_\varepsilon}\in \Phi_{n,2d}\subset
D_{\sigma_0,\sigma,2d,n^d}$ for  sufficient large $\mathcal
B_n,\mathcal C_n$ and $\Xi_n$. To present the covering number of
$\Phi_{n,2d}$, we need the
 following smoothness assumption on  $\sigma$.

\begin{assumption}\label{Ass:sigmoidal}
$\sigma$ is a non-decreasing sigmoidal function satisfying
\begin{equation}\label{lip for sigma}
       |\sigma(t)-\sigma(t')|\leq C_\sigma|t-t'|.
\end{equation}
\end{assumption}

Assumption \ref{Ass:sigmoidal} has already been adopted in
\cite[Theorem 5.1]{Kurkova2007} and \cite[Lemma 2]{Makovoz1996} to
quantify the covering number of some shallow nets. It should be
mentioned that there are numerous functions satisfying Assumption
\ref{Ass:sigmoidal}, including the widely used functions presented
in Figure 4. With these helps, we present a tight estimate  for the
covering number of $\Phi_{n,2d}$ in the following proposition, whose
proof will be given in Appendix B.
\begin{figure}[!t]
\begin{minipage}[b]{.49\linewidth}
\centering
\includegraphics*[scale=0.30]{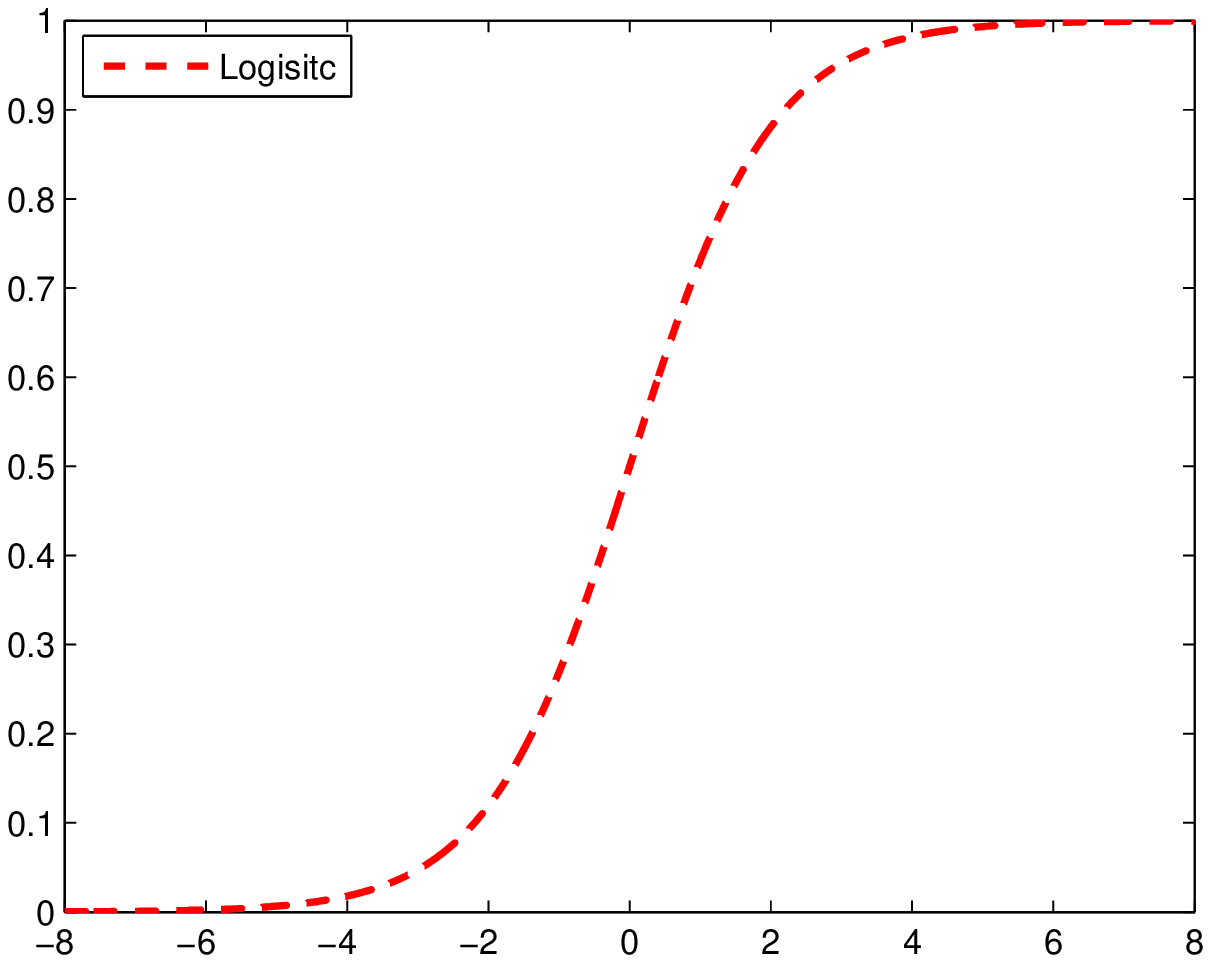}
\centerline{{\small (a) $
               \sigma(t)=\frac{1}{1+e^{-t}},
$}}
\end{minipage}
\hfill
\begin{minipage}[b]{.49\linewidth}
\centering
\includegraphics*[scale=0.30]{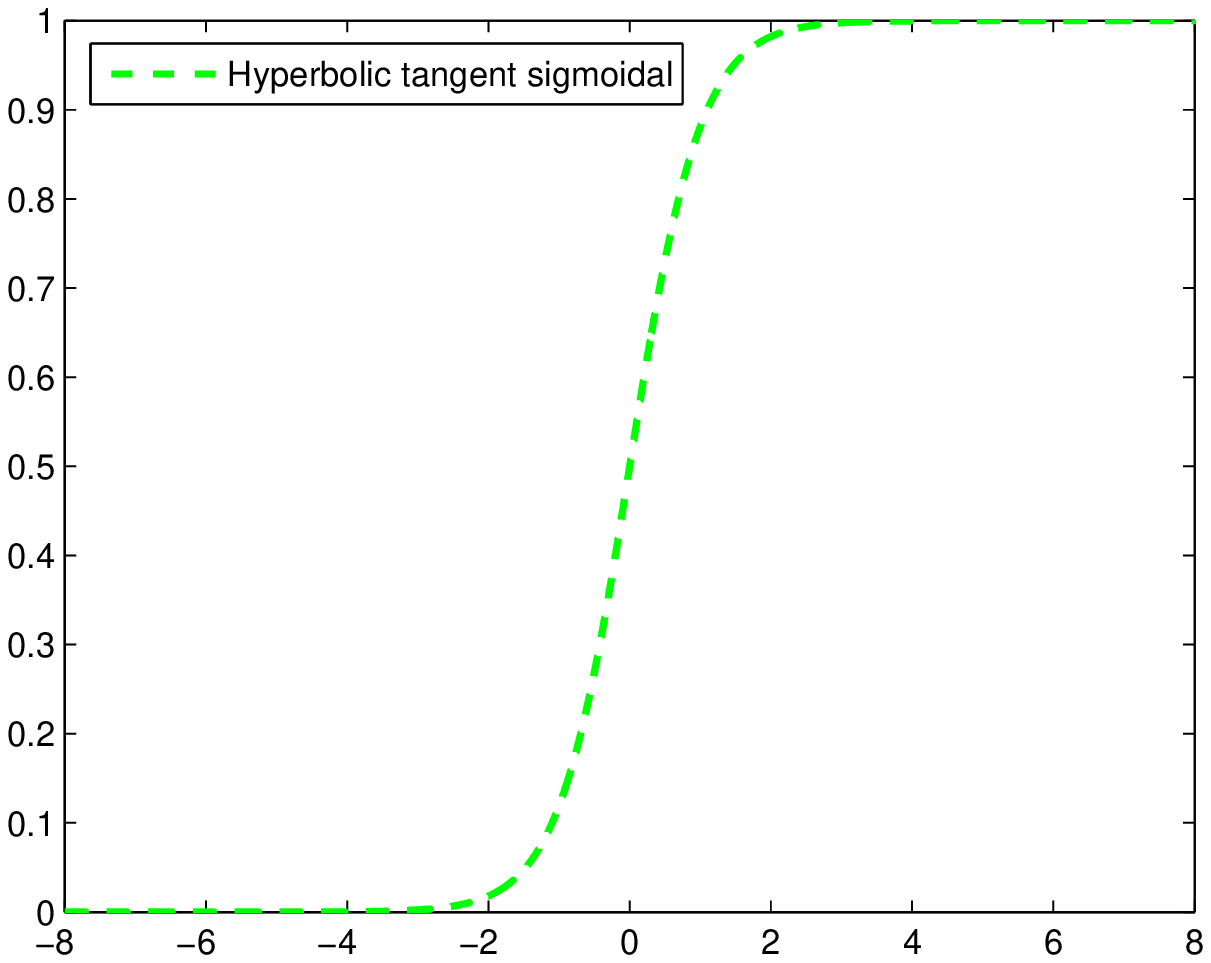}
\centerline{{\small (b) $\sigma(t)=\frac12(\tanh(t)+1)$}}
\end{minipage}
\begin{minipage}[b]{.49\linewidth}
\centering
\includegraphics*[scale=0.30]{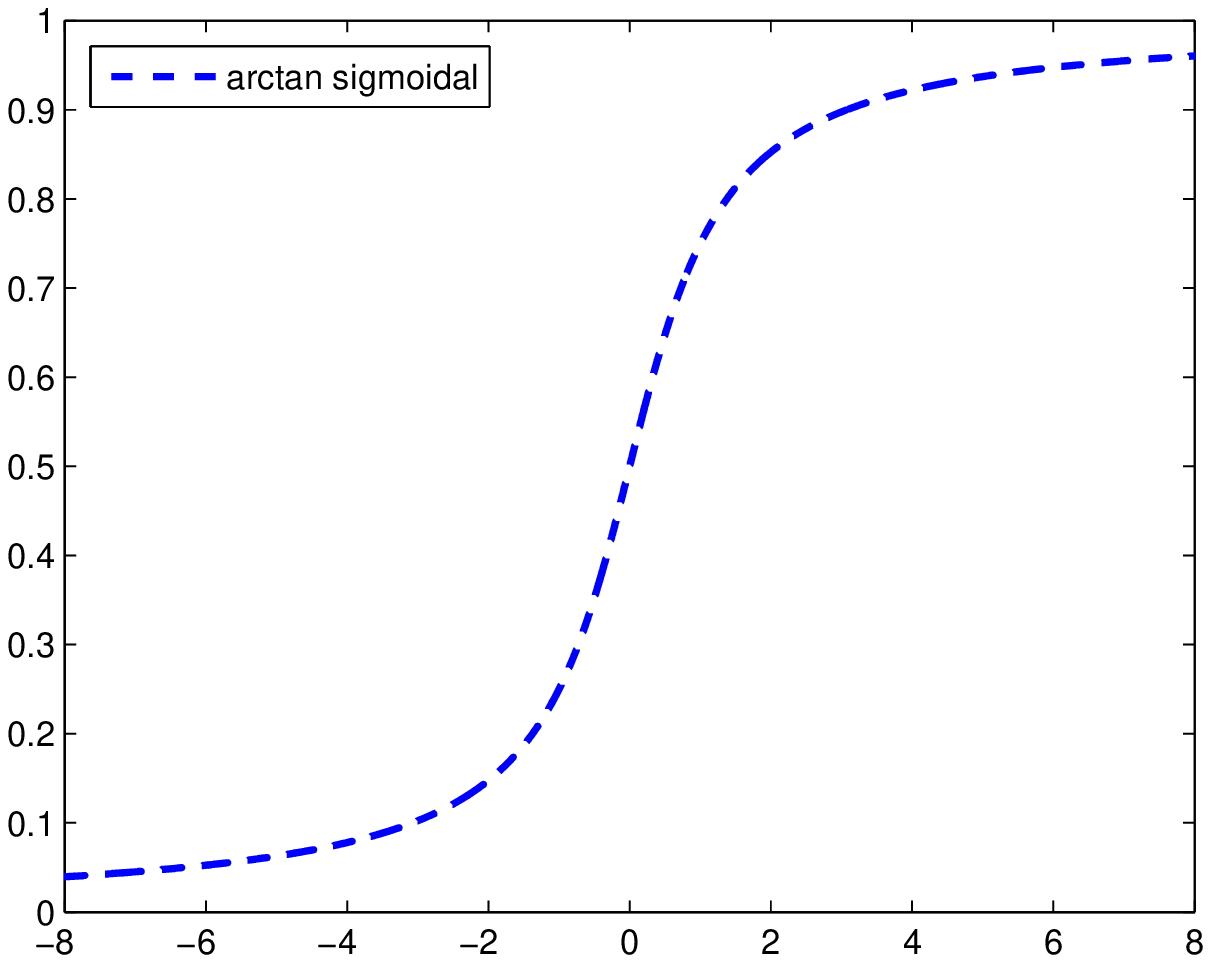}
\centerline{{\small (c) $\sigma(t)=\frac1{\pi}\arctan(t)+\frac12$}}
\end{minipage}
\hfill
\begin{minipage}[b]{.49\linewidth}
\centering
\includegraphics*[scale=0.30]{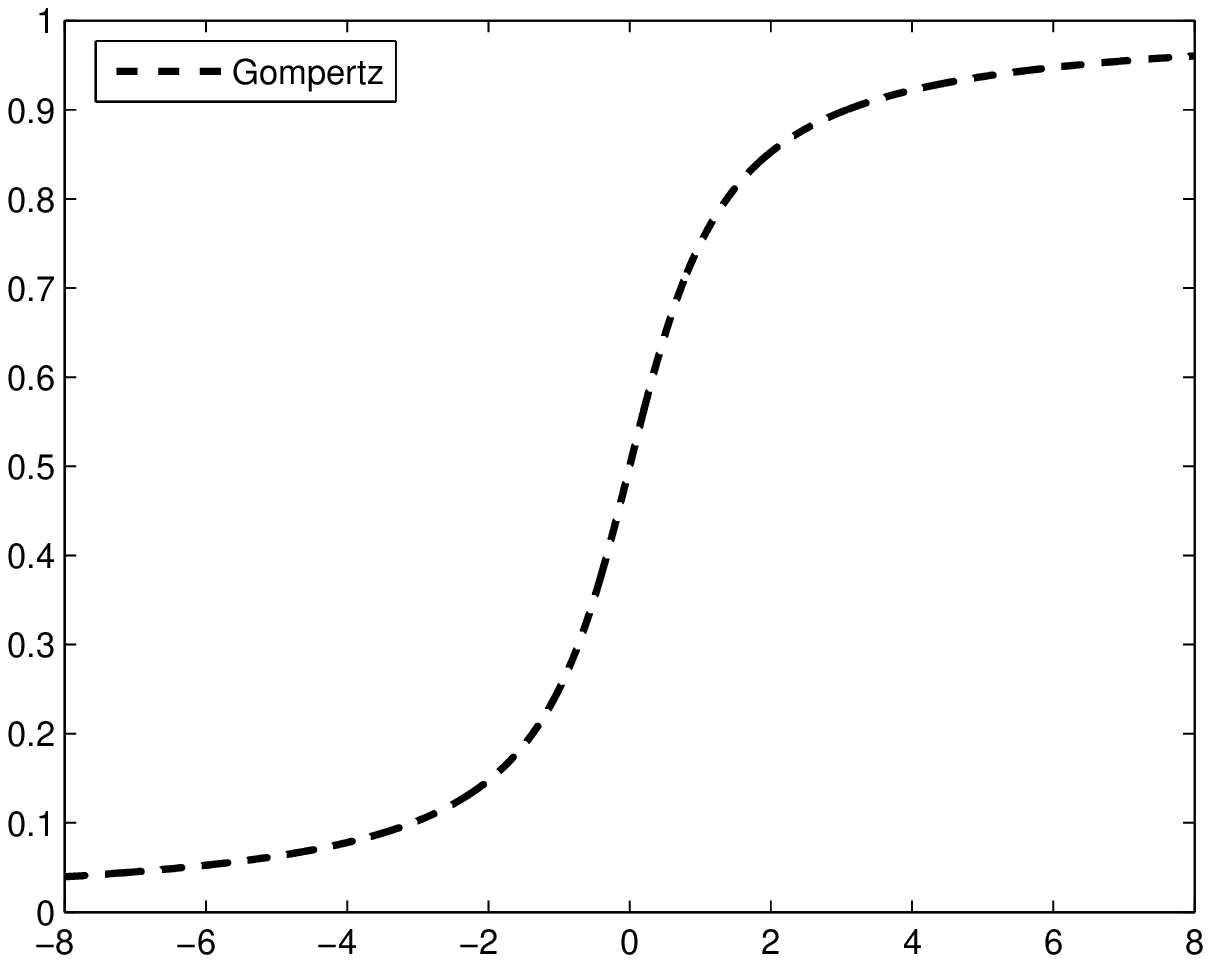}
\centerline{{\small (d) $\sigma(t)=e^{- e^{-t}}$}}
\end{minipage}
\hfill \caption{Four widely used activation functions: (a) logistic
function; (b) hyperbolic tangent function; (c) arctan sigmoidal
function; (d)  Gompertz function}
\end{figure}

\begin{proposition}\label{Proposition:covering number}
Let $\Phi_{n,2d}$ be defined by (\ref{Hypothesis space}). Under
Assumption  \ref{Ass:sigmoidal}, there holds
\begin{eqnarray*}
     &&\log\mathcal N(\varepsilon, \Phi_{n,2d})
      \leq 4dn^d\log\log\frac{3e (2d+1)\mathcal C_nC_\sigma
      n^{d}\Xi_n }{\varepsilon}
      \\
     &+&
   n^d\log \frac{4\mathcal
      B_n(24e^2)^{2d}(2d+1)^{6d}
      \mathcal C_n^{6d+2}\Xi_n^{6d}C_\sigma^{6d+1}
      n^{6d^2+2d}}{\varepsilon^{6d+2}}.
 \end{eqnarray*}
\end{proposition}

In \cite{Makovoz1996,Kurkova2007}, a   bound of the covering number
for the set
$$
    \mathcal F:= \{f=\sigma(wx+b):w\in\mathbb R^d, b\in\mathbb R, \|f\|_*\leq 1\}
$$
with $\|\cdot\|_*$ denoting some norm including the uniform norm and
$\sigma$ satisfying Assumption \ref{Ass:sigmoidal}  was derived.  It
is obvious that $ \mathcal F$ is a shallow net of only one neurons.
Based on this interesting result, \cite[Chap.16]{Gyorfi2002} and
\cite{Maiorov2006b} presented a tight estimate for $\mathcal
N(\varepsilon, S^*_{\sigma,n})$ as
\begin{equation}\label{covering for shallow}
        \mathcal N(\varepsilon, S^*_{\sigma,n})=\mathcal O\left(n^d
        \log\frac{\Gamma_n}{\varepsilon}\right),
\end{equation}
where
$$
        S^*_{\sigma,n}:=\left\{f=\sum_{j=1}^nc_j\sigma(w_jx+\theta_j):
        |c_j|\leq\Gamma_n, w_j,\theta_j\in\mathbb R\right\}
$$
and $\Gamma_n>0$ and $\sigma$ satisfies (\ref{Ass:sigmoidal}). Here,
we should highlight that the bounded assumption $|c_j|\leq \Gamma_n$
for the outer weights  are necessary, without which the capacity
should be infinity according to theory of
\cite{Maiorov1999b,Maiorov2006b}.

If  $\mathcal B_n$, $\mathcal C_n$, $\Xi_n$ and $\Gamma_n$ are not
very large, i.e., do not grow  exponentially with respect to $n$,
then it follows from Proposition \ref{Proposition:covering number}
that
\begin{equation}\label{Covering number of simplfied}
       \log\mathcal N(\varepsilon, \Phi_{n,2d})
       =
       \mathcal O\left(n^d \log\frac{n}{\varepsilon}\right),
\end{equation}
which is the same as (\ref{covering for shallow}). Comparing
$\Phi_{n,2d}$ with $S^*_{\sigma,n}$, we find that adding a layer
with bounded parameters does not enlarge the covering number.
 Thus, Proposition
\ref{Proposition:covering number} together with Proposition
\ref{Proposition:localization} yields that deep nets can approximate
more functions than shallow nets without increasing the covering
number of shallow nets. Proposition \ref{Proposition:covering
number} and Proposition \ref{Proposition:sparse approximation} show
that deep nets can approximate sparse function better than shallow
nets within the same price.

\section{Learning Rate Analysis}\label{Sec.Main result}
In this section, we present the ERM algorithm on deep nets and
provide its near optimal learning rates in learning Lipschitz
functions and sparse functions in the framework of learning theory
\cite{Cucker2007}.

\subsection{Algorithm and assumptions}

In learning theory \cite{Cucker2007},  samples
$D_m=(x_i,y_i)_{i=1}^m$ are assumed to be drawn independently
 according to $\rho$,  a Borel probability measure  on ${\mathcal Z}
 ={\mathcal X}\times {\mathcal Y}$ with $\mathcal
X=\mathbb I^d$ and $\mathcal Y\subseteq[-M,M]$ for some $M>0$. The
primary objective is the regression function defined by
$$
         f_\rho(x)=\int_{\mathcal Y} y d\rho(y|x), \qquad x\in\mathcal
         X
$$
which minimizes the generalization error
$$
         \mathcal E(f):=\int_{\mathcal Z}(f(x)-y)^2d\rho,
$$
where $\rho(y|x)$ denotes the conditional distribution at $x$
induced by $\rho$.  Let $\rho_X$ be the marginal distribution of
$\rho$ on $\mathcal X$ and $(L^2_{\rho_{_X}}, \|\cdot\|_\rho)$ be
the Hilbert space of $\rho_X$ square integrable functions on
$\mathcal X$. Then   for arbitrary $f\in L_{\rho_X}^2$, there holds
\cite{Cucker2007}
\begin{equation}\label{equality}
        \mathcal E(f)-\mathcal E(f_\rho)=\|f-f_\rho\|_\rho^2.
\end{equation}

We devote to deriving learning rate for the following ERM algorithm
\begin{equation}\label{ERM}
      f_{D,n}:=\arg\min_{f\in\Phi_{n,2d}}\frac1m\sum_{i=1}^m[f(x_i)-y_i]^2,
\end{equation}
where $\Phi_{n,2d}$ is the set of deep nets  defined by
(\ref{Hypothesis space}). Before presenting the main results, we
should introduce some assumptions.

\begin{assumption}\label{Ass:regression}
We assume $f_\rho\in Lip^{(r,c_0)}$.
\end{assumption}

Assumption \ref{Ass:regression} is the $r$-Lipschitz continuous
condition for the regression function, which is  standard  in
learning theory
\cite{Gyorfi2002,Kohler2005,Maiorov2006a,Cucker2007,Lin2015,Lin2017}.
To show the advantage of deep nets learning, we should add the
sparseness assumption on $f_\rho$.

\begin{assumption}\label{Ass:regression1}
We assume $f_\rho\in Lip^{(N,s,r,c_0)}$.
\end{assumption}

Assumption \ref{Ass:regression1} shows that $f_\rho$ is $s$-sparse
in $N^d$ partitions.  The additional sparseness assumption is
natural in applications like image processing \cite{Wang2004} and
computer vision \cite{Bradski2008}.

\begin{assumption}\label{Ass:margin}
There exists some constant  $c_1>0$ such that $\|f\|_\rho\leq
c_1\|f\|_{L^2(\mathbb I^d)}$.
\end{assumption}

Assumption \ref{Ass:margin} concerns the distortion of the marginal
distribution $\rho_X$. It has been utilized in \cite{Zhou2006} and
\cite{Shi2013} to quantify the learning rates of support vector
machines and kernel lasso. It is obvious that this assumption holds
for the uniform distribution. If $f_\rho$ is supported on $S$ but
$\rho_X$ is supported on $\mathbb I^d\backslash S$, it is impossible
to derive a satisfactory learning rate. Thus, Assumption
\ref{Ass:margin} is important and necessary to show the sparseness
of $f_\rho$ in the spacial domain. Let $\mathcal M$ be the class of
all Borel measures $\rho$ on $\mathcal Z$ satisfying Assumption
\ref{Ass:regression}.
Let $\mathbf G_m$ be the set of
all estimators derived from the samples $D_m$. Define
$$
          e_m(\Theta):=\inf_{f_D\in\mathbf G_m}\sup_{\rho\in
          \Theta}\mathbf E\left\{\|f_\rho-f_{D}\|^2_\rho\right\}.
$$
 Then
it can be found in \cite[Theorem 3.2]{Gyorfi2002}
that
\begin{equation}\label{baseline}
            e_m(\mathcal M)\geq \tilde{C}m^{-\frac{2r}{2r+d}},\ m=1,2,\dots,
\end{equation}
where  $\tilde{C}$ is a constant depending only on $c_0$, $c_1$,
$M$, $r$ and $d$.

\begin{assumption}\label{Ass:space}
Let $\Xi_n\geq 2L$, $\mathcal B_n\geq 2d$  and $\mathcal C_n\geq M$,
where $L$ satisfies
\begin{equation}\label{definition K for sigmoidal 222}
                 \left\{\begin{array}{cc}
                 |\sigma(t)-1|<n^{-r-d}\left(\frac{s}{N^d}\right)^{\frac12},&
                 \mbox{if}\
                 t\geq L, \\
                 |\sigma(t)|<n^{-r-d}\left(\frac{s}{N^d}\right)^{\frac12},
                &
                \mbox{if}\ t\leq-L.\end{array}\right.
\end{equation}
\end{assumption}

It is obvious that  $L$ depends only on $\sigma$, $s$, $N$ and $n$.
Assumption \ref{Ass:space} is technical and describes the capacity
of $\Phi_{n,2d}$. It guarantees that the space $\Phi_{n,2d}$ is
large enough to contain $N^*_{n,{\bf j},L}$.   Furthermore, the
solvability of (\ref{ERM}) depends heavily on the concrete values
$\mathcal B_n$, $\mathcal C_n$ and $\Xi_n$  \cite{Gyorfi2002}.

\subsection{Learning rate analysis}

Since $|y|\le M$ almost everywhere, we have $|f_\rho(x)|\leq M$. It
is   natural for us to project an output function $f: {\mathcal X}
\rightarrow\mathbb R$ onto the interval $[-M, M]$ by the projection
operator
$$
          \pi_M f(x):=\left\{\begin{array}{ll}
         f(x), & \mbox{if}\ -M\leq f(x)\leq M,\\
         M, & \mbox{if}\ f(x)>M,\\
         -M, &\mbox{if}\ f(x)<-M. \end{array}\right.
$$
Thus, the estimate we studied in this paper is $\pi_Mf_{D,n}$.

 The main results of this paper are the following
two learning rate estimates. In the first one, we present the
learning rate for algorithm (\ref{ERM}) when the smoothness
information of the regression function is given.

\begin{theorem}\label{Theorem: ERM}
Let $0<\delta<1$ and $f_{D,n}$ be defined by (\ref{ERM}). Under
Assumptions \ref{Ass:sigmoidal}, \ref{Ass:regression} and
\ref{Ass:space}, if $n=\left\lfloor
m^{\frac{d}{2s+d}}\right\rfloor$, then with confidence at least
$1-\delta$, there holds
\begin{equation}\label{theorem1}
      \mathcal E(\pi_Mf_{D,n})-\mathcal E(f_\rho)
      \leq
      Cm^\frac{-2r}{2r+d}\log\left(\mathcal
       B_n
      \mathcal C_n\Xi_nm\right)\log\frac2\delta,
\end{equation}
where $C$ is a constant independent of $\delta$, $n$ or $m$.
\end{theorem}

From Theorem \ref{Theorem: ERM}, we can derive the following
corollary, which states the near optimality of the derived learning
rate for $\pi_Mf_{D,n}$.

\begin{corollary}\label{Corollary: optimal}
Under Assumptions \ref{Ass:sigmoidal}, \ref{Ass:regression}  and
\ref{Ass:space}, if $n=\left\lfloor
m^{\frac{d}{2r+d}}\right\rfloor$, then
\begin{eqnarray*}
     &&\tilde{C}m^{-\frac{2r}{2r+d}}\leq \max_{f_\rho\in Lip^{(r,C_0)}}\mathbf E\left[\mathcal E(\pi_Mf_{D,n})-\mathcal
      E(f_\rho)\right]\\
     &\leq&
     (2+\log2) C m^{-\frac{ 2r}{2r+d}}\log( \mathcal
       B_n
      \mathcal C_n\Xi_nm).
\end{eqnarray*}
\end{corollary}

The proofs of Theorem \ref{Theorem: ERM} and Corollary
\ref{Corollary: optimal} will be postponed to Appendix C. It is
shown in Theorem \ref{Theorem: ERM} and Corollary \ref{Corollary:
optimal} that implementing ERM on $\Phi_{n,2d}$ can reach the near
optimal learning rates (up to a logarithmic factor) provided
$\Xi_n$, $\mathcal C_n$  and $\mathcal B_n$ are not very large. In
fact, neglecting the solvability of algorithm (\ref{ERM}), we can
set $\mathcal B_n=2d$, $\mathcal C_n=M$ and $\Xi_n=2L$. Due to
(\ref{definition K for sigmoidal 222}), the concrete value of $L$
depends on $\sigma$. Taking the logistic function for example, we
can set $L=(r+d)\log (nN^d/s)$. Theorem \ref{Theorem: ERM} and
Corollary \ref{Corollary: optimal} yield that for some easy learning
task (exploring only the smoothness information of $f_\rho$), deep
nets perform at least not worse than shallow nets and can reach the
almost optimal learning rates for all learning schemes.

In the following theorem, we show that for some difficult learning
task (exploring sparseness and smoothness information of $f_\rho$),
deep nets learning can break through the bottleneck of shallow nets
learning via establishing a learning rate much faster than
(\ref{baseline}).

\begin{theorem}\label{Theorem: ERM sparse}
Let $0<\delta<1$ and $f_{D,n}$ be defined by (\ref{ERM}). Under
Assumptions \ref{Ass:sigmoidal}, \ref{Ass:regression1},
\ref{Ass:margin} and \ref{Ass:space}, if $n=\left\lfloor
\left(\frac{ms}{N^d}\right)^{\frac{d}{2r+d}}\right\rfloor$ and
$m\geq \frac{4^{2r+d}N^{2r+2d}}{s}$, then with confidence at least
$1-\delta$, there holds
\begin{eqnarray}\label{theorem sparse}
      &&\mathcal E(\pi_Mf_{D,n})-\mathcal E(f_\rho)\nonumber\\
      &\leq&
      C'm^{-\frac{2r}{2r+d}}\log\left(\mathcal
       B_n
      \mathcal C_n\Xi_nm\right)\left(\frac{s}{N^d}\right)^{\frac{d}{2r+d}}\log\frac2\delta,
\end{eqnarray}
where $C'$ is a constant    independent of $N$, $s$, $\delta$, $n$
or $m$.
\end{theorem}

Similarly, we can obtain the following corollary, which exhibits the
 derived learning rate in expectation.

\begin{corollary}\label{Corollary: optimal sparse}
Under Assumptions \ref{Ass:sigmoidal}, \ref{Ass:regression1},
\ref{Ass:margin} and \ref{Ass:space}, if $n=\left\lfloor
\left(\frac{ms}{N^d}\right)^{\frac{d}{2s+d}}\right\rfloor$ and
$m\geq \frac{4^{2r+d}N^{2r+2d}}{s}$, then
\begin{eqnarray*}
      &&\mathbf E\left[\mathcal E(\pi_Mf_{D,n})-\mathcal
      E(f_\rho)\right]\\
     &\leq&
     (2+\log2) C' m^\frac{-2r}{2r+d}\log( \mathcal
       B_n
      \mathcal C_n\Xi_nm)\left(\frac{s}{N^d}\right)^{\frac{d}{2r+d}}.
\end{eqnarray*}
\end{corollary}

Theorem \ref{Theorem: ERM sparse} and Corollary \ref{Theorem: ERM
sparse}, whose proofs will be given in Appendix C,  show that if the
additional sparseness information is imposed, then ERM based on deep
nets can break through the optimal learning rates   in
(\ref{baseline}) for shallow nets. To be detailed, if $f_\rho$ is
1-sparse in $ m^{\frac1{2r+2d}}$ partitions, then we can take
$\sigma$ be the logistic function and $\mathcal B_n=2d$, $\mathcal
C_n=M$ and $\Xi_n=2(r+d)\log (nN^d)$ to get a learning rate of order
$m^{-\frac{2r}{2r+d}-\frac{d}{2r+2d}}\ll m^{-\frac{2r}{2r+d}}.$ This
 shows the advantage of deep nets in learning sparse
functions.

\section{Related Work and Discussions}\label{Sec.Related work}
Stimulated by the great success of deep learning in applications
understanding deep learning as well as its theoretical verification
becomes a hot topic in approximation and statistical learning
theory. Roughly speaking, the studies of deep net approximation can
be divided into two categories:  deducing the limitations of shallow
nets and pursuing the advantages of deep nets.

Limitations of the approximation capabilities of shallow nets were
firstly proposed in \cite{Blum1991} in terms of  their incapability
of localized approximation. Five years later,  \cite{Chui1996}
described their limitations via providing lower bounds of
approximation of smooth functions in the minimax sense, which was
recently highlighted by \cite{Lin2017a} via showing that  there
exists a probabilistic measure, under which, all smooth functions
cannot be approximated by shallow nets very well with high
confidence.  In  \cite{Bengio2006}, Bengio et al. also  pointed out
the limitations of some shallow nets in terms of the so-called
``curse of dimensionality''.  In  some recent interesting papers
\cite{Kurkova2017,Kurkova2017,Kurkova2018}, limitations of shallow
nets were  presented in terms of establishing lower bound of
approximating functions with different variation restrictions.

Studying advantages of deep nets is also a classical topic in neural
networks approximation. It can date back to 1994, where Chui et al.
\cite{Chui1994} deduced the localized approximation property of deep
nets which is far beyond the capability of shallow nets
\cite{Blum1991}. Recently, more and more advantages of deep nets
were theoretical verified in the approximation theory community. In
particular,
  \cite{Mhaskar2016} showed the power of depth
of neural network in approximating   hierarchical functions;
 \cite{Shaham2015} demonstrated that deep nets can improve the
approximation capability of shallow nets when the data are located
on a manifold;   \cite{LinH2017} presented the necessity of deep
nets in physical problems which possess symmetry, locality or
sparsity; \cite{McCane2017} exhibited the outperformance of deep
nets in approximating radial functions and so on. Compared with
these results, we focus on show the good performance of deep nets in
approximation sparse functions in the spacial domain and also study
the cost for the approximation, just as Propositions
\ref{Proposition:sparse approximation} and \ref{Proposition:covering
number} exhibited.

In the learning theory community,  learning rates for ERM on shallow
nets with certain activation functions were studied in
\cite{Maiorov2006a}. Under Assumption \ref{Ass:regression},
\cite{Maiorov2006a} derived a near optimal learning rate of order $
m^\frac{-2r}{2r+d}\log^2 m$. The novelty of our Theorem
\ref{Theorem: ERM} is that we focus on learning rates of ERM on deep
nets rather than shallow nets, since deep nets studied in this paper
can provide localized  approximation. Our result together with
\cite{Maiorov2006a} demonstrates that deep nets can learn more
functions (such as the indicator function) than shallow nets without
sacrificing the generalization capability of shallow nets. However,
since deep nets possess the sparse approximation property, it is
stated in Theorem \ref{Theorem: ERM sparse} that if additional
a-priori information is given, then deep nets can breakthrough the
optimal learning rate for shallow nets, showing the power of depth
in neural networks learning.  Learning rates for shallow nets
equipped with a so-called complexity penalization strategy were
presented in \cite[Chapter 16]{Gyorfi2002}. However, only variance
estimate rather than the learning rate were established in
\cite{Gyorfi2002}. More importantly, their  algorithms and network
architectures  are different from our paper.

  In the recent work
\cite{Lin2016},   a   neural network  with  two hidden layers was
developed  for the learning purpose and   the optimal learning rates
of order $ m^\frac{-2r}{2r+d}$ were presented. It should be noticed
that the main idea of the construction in \cite{Lin2016} is the
local average argument rather than  any optimization strategy such
as (\ref{ERM}). Furthermore, \cite{Lin2016}'s network architecture
is a hybrid of feed-forward neural network (second hidden layer) and
radial basis function networks (first hidden layer). The constructed
network in the present paper is a standard deep net possessing the
same network architectures  in both hidden layers.

In our previous work \cite{Chui2016a}, we constructed a deep net
with three hidden layers when $\mathcal X$ is in a $d^*<d$
dimensional sub-manifold and provided a learning rate of order
$m^{-\frac{2r}{2r+d^*}}$. The construction in \cite{Chui2016a} were
based on the local average argument \cite{Gyorfi2002}. The main
difference between the present paper and \cite{Chui2016a} is that we
used the optimization strategy in determining the parameters of deep
nets rather than construct them directly.   In particular, the main
tool in this paper is a refined estimate for the covering number.

Another  related work is \cite{Kohler2005}, which provided error
analysis of a complexity regularization scheme whose hypothesis
space is deep nets with two hidden layers proposed in
\cite{Mhaskar1993}. They derived a learning rate of  $\mathcal
O(m^{-2r/(2r+D)}(\log m)^{4r/(2r+d)})$ under Assumption
\ref{Ass:regression}, which is the same as the rate in Theorem
\ref{Theorem: ERM} up to a logarithmic factor. Neglecting the
 algorithmic factor, the main novelty of our work is that our analysis
combines the expressivity (localized approximation) and
generalization capability, while \cite{Kohler2005}'s result concerns
only  the generalization capability. We refer the readers to
\cite{Bottou1992,Chui1994} for some advantages of   localized
approximation and sparse approximation in the spacial domain.

To finalize the discussion, we mention that the present paper only
compares deep nets with two hidden layers with shallow nets and
demonstrates the advantage  of the former architecture from
approximation learning theory viewpoints. As far as the optimal
learning rate is concerned, to theoretically provide the power of
depth,  more restrictions on the regression function should be
imposed. For example, shallow nets are capable of exploring the
smoothness information \cite{Maiorov2006a}, deep nets with two
hidden layers can tackle both sparseness and smoothness information
(Theorem \ref{Theorem: ERM sparse} in this paper), and deep nets
with more hidden layers succeed in handling sparseness information,
smoothness information and manifold features of the input space
(combining Theorem \ref{Theorem: ERM sparse} in this paper with
Theorem 1 in \cite{Chui2016a}). In a word, deep nets with more
hidden layers can embody more information for the learning task. It
is interesting to study the power of depth along such flavor and
determine which information can (or cannot) be explored by deepening
the networks.

\section{Conclusion}
In this paper, we analyzed the expressivity and generalization of
deep nets. Our results showed that without essentially enlarging the
capacity of shallow nets, deep nets possess excellent expressive
power in terms of providing localized approximation and sparse
approximation. Consequently, we proved that for some difficult
learning tasks (exploring both sparsity and smoothness), deep nets
could break though the optimal learning rates established for
shallow nets. All these results showed the power of depth from the
  learning theory viewpoint.

\section*{Appendix A: Proofs of Propositions \ref{Proposition:localization} and \ref{Proposition:sparse approximation}}
In this Appendix, we present the proofs of Propositions
\ref{Proposition:localization} and \ref{Proposition:sparse
approximation}. The basic idea  of our proof was motivated by
\cite{Chui1994} and the property (\ref{definition K for sigmoidal
1}) of    sigmoidal functions.

\begin{IEEEproof}[Proof of Proposition
\ref{Proposition:localization}]  When $x\notin A_{n, {\bf j}}$,
there exists an $\ell_0$ such that $
       |x^{(\ell_0)}-\xi_{\bf j}^{(\ell_0)}|>\frac1{2n}. $ If $
     x^{(\ell_0)}-\xi_{\bf j}^{(\ell_0)}<-1/(2n),
$ then
$$
    1/(2n)+x^{(\ell_0)}-\xi_{\bf j}^{(\ell_0)}< 0.
$$
If
    $x^{(\ell_0)}-\xi_{\bf j}^{(\ell_0)}>1/(2n),$ then
$$
   1/(2n)-x^{(\ell_0)}+\xi_{\bf j}^{(\ell_0)}< 0.
$$
The above assertions together with the definition of $\sigma_0$
yield
$$
       \sum_{\ell=1}^d\sigma_0\left[\frac1{2n}+x^{(\ell)}-\xi_{\bf j}^{(\ell)}\right]
      +\sum_{\ell=1}^d\sigma_0\left[\frac1{2n}-x^{(\ell)}+\xi_{\bf j}^{(\ell)}\right]
      < 2d-1.
$$
Thus,
\begin{eqnarray*}
       &&\sum_{\ell=1}^d\sigma_0\left[\frac1{2n}+x^{(\ell)}-\xi_{\bf j}^{(\ell)}\right]
      +\sum_{\ell=1}^d\sigma_0\left[\frac1{2n}-x^{(\ell)}+\xi_{\bf
      j}^{(\ell)}\right]\\
      &&-2d+1/2<-1/2,
\end{eqnarray*}
which together with (\ref{definition K for sigmoidal 1}) and
(\ref{NN for localization}) yields
$$
       |N^*_{n,{\bf j},K_\varepsilon}(x)|<\varepsilon.
$$
This finishes the proof of part (a).  We turn to prove  assertion
(b) in Proposition \ref{Proposition:localization}. Since $x\in A_{n,
{\bf j}}$, for all $1\leq \ell\leq d$, there holds $
|x^{(\ell)}-\xi_{\bf j}^{(\ell)}|\leq \frac1{2n}$. Thus, for all
$\xi\in
  A_{n, {\bf j}}$, there holds
$$
       \frac1{2n}\pm(x^{(\ell)}-\xi_{\bf j}^{(\ell)})\geq 0.
$$
It follows from the definition of $\sigma_0$ that
$$
       \sum_{\ell=1}^d\sigma_0\left[\frac1{2n}+x^{(\ell)}-\xi_{\bf j}^{(\ell)}\right]
      +\sum_{\ell=1}^d\sigma_0\left[\frac1{2n}-x^{(\ell)}+\xi_{\bf j}^{(\ell)}\right]
      =2d.
$$
That is,
\begin{eqnarray*}
       &&\sum_{\ell=1}^d\sigma_0\left[\frac1{2n}+x^{(\ell)}-\xi_{\bf j}^{(\ell)}\right]
      +\sum_{\ell=1}^d\sigma_0\left[\frac1{2n}-x^{(\ell)}+\xi_{\bf
      j}^{(\ell)}\right]\\
     && -2d+1/2=1/2.
\end{eqnarray*}
Hence, (\ref{definition K for sigmoidal 1}) implies
$$
     |N^*_{n,{\bf j},K_\varepsilon}(x)-1|<\varepsilon.
$$
Since $\sigma$ is non-decreasing, we have $N^*_{n,{\bf j},K}(x)\leq
1$ for all $x\in[0,1]^d$.  The proof of Proposition
\ref{Proposition:localization} is finished.
\end{IEEEproof}

\begin{IEEEproof}[Proof of Proposition \ref{Proposition:sparse approximation}]
 Since $\mathbb I^d=\bigcup_{{\bf j}\in\mathbb N_n^d}A_{n,{\bf j}}$, for each
$x\in\mathbb I^d$, there exists a  ${\bf j}_x$ such that $x\in
A_{n,{\bf j}_x}$. Here, if $x$ lies on the boundary of some
$A_{n,{\bf j}}$, we denote by ${\bf j}_x$  an arbitrary but fixed
${\bf j}$ satisfying $A_{n,{\bf j}}\ni x$. Then, it follows from
(\ref{Def.N2}) that
\begin{eqnarray}\label{sparse decomposition}
   f(x)- N_{n,\eta,K_\varepsilon}(x)
   &=&
   f(x)-f(\eta_{{\bf j}_x})-\sum_{{\bf j}\neq{\bf j}_x}f(\eta_{\bf j})N^*_{n,{\bf
   j},K_\varepsilon}(x) \nonumber\\
   &+&
   f(\eta_{{\bf j}_x})[1-N^*_{n,{\bf
   j}_x,K_\varepsilon}(x)].
\end{eqnarray}
We get from (\ref{sparse decomposition}), (\ref{lip}), $x,\eta_{{\bf
   j}_x}\in A_{n,{\bf j}_x}$ and Proposition \ref{Proposition:localization} that
\begin{eqnarray*}
   &&|f(x)- N_{n,\eta,K_\varepsilon}(x)|\\
   &\leq& c_0\|x-\eta_{{\bf
   j}_x}\|^r+(n^d-1)\|f\|_{L^\infty(\mathbb
    I^d)}\varepsilon
     + \|f\|_{L^\infty(\mathbb
    I^d)}\varepsilon\\
     &\leq&
    2^{r/2}c_0n^{-r}+n^d\|f\|_{L^\infty(\mathbb
    I^d)}\varepsilon.
\end{eqnarray*}
This proves (\ref{sparse approximation1}). If $x\notin\bigcup_{{\bf
k}\in\Lambda_s}\overline{\Lambda_{\bf k}}$, then $A_{n,{\bf
j}_x}\cap S=\varnothing.$ Thus, for arbitrary   $\eta_{{\bf j}_x}$
satisfying $\eta_{{\bf j}_x}\in A_{n,{\bf j}_x}$, we have from
$f_\rho\in Lip^{(N,s,r,c_0)}$, Proposition
\ref{Proposition:localization} and (\ref{Def.N2}) that
\begin{eqnarray*}
     &&|N_{n,\eta,K_\varepsilon}(x)|\\
    &=&
     \sum_{{\bf j}\neq{\bf j}_x}f(\eta_{\bf j})N^*_{n,{\bf
   j},K_\varepsilon}(x)
    +
   f(\eta_{{\bf j}_x}) N^*_{n,{\bf
   j}_x,K_\varepsilon}(x)\\
   &\leq&
   \|f\|_{L^\infty(\mathbb I^d)}\sum_{{\bf j}\neq{\bf j}_x}N^*_{n,{\bf
   j},K_\varepsilon}(x)
   \leq
   \|f\|_{L^\infty(\mathbb I^d)}n^d\varepsilon.
\end{eqnarray*}
This proves (\ref{sparse approximation2}) and completes the proof of
Proposition \ref{Proposition:sparse approximation}.
\end{IEEEproof}

\section*{Appendix B: Proofs of Proposition \ref{Proposition:covering
number}}

The aim of this appendix is to prove Proposition
\ref{Proposition:covering number}. Our main idea is to decouple
different hidden layers by using Assumption \ref{Ass:sigmoidal} and
the definition of the covering number. For this purpose, we need the
following five lemmas. the first two can be found in \cite[Lemma
16.3]{Gyorfi2002} and \cite[Theorem 9.5]{Gyorfi2002}, respectively.
The third one can be easily deuced from \cite[Lemma
9.2]{Gyorfi2002}, \cite[Theorem 9.4]{Gyorfi2002} with $p=1$ and the
fact $\mathcal N(\varepsilon,\mathcal F)\leq \mathcal N(\varepsilon,
\mathcal F,L^1(\mathcal X))$. The last two are well known, and we
present their proofs for the sake of completeness.

\begin{lemma}\label{Lemma:relation vc dimension}
Let $\mathcal F$ be a family of real functions and let $h:\mathbb
R\rightarrow \mathbb R$ be a fixed nondecreasing function. Define
the class $\mathcal G=\{h\circ f:f\in\mathcal F\}$. Then
$$
        V_{\mathcal G^+}\leq V_{\mathcal F^+},
$$
where
$$
   \mathcal H^+:=\left\{\{(z,t)\in\mathbb R^d\times\mathbb R;t\leq
   h(z)\}:h\in\mathcal H\right\}
$$
for some set of functions $\mathcal H$ and $V_U$ denotes the VC
dimension \cite{Gyorfi2002} of the set $U$ over $\mathcal X$.
\end{lemma}

\begin{lemma}\label{Lemma:esitmate vc dimenision}
Let $\mathcal G$ be an $r$-dimensional vector space of real
functions on $\mathbb R^d,$ and set
$$
           \mathcal A=\left\{\{z:g(z)\geq0\}:g\in\mathcal G\right\}.
$$
Then
$$
         V_\mathcal A\leq r.
$$
\end{lemma}

\begin{lemma}\label{lemma:inner covering number}
Let $\mathcal F$ be a class of functions $f:\mathbb
R^d\rightarrow[0,M^*]$ with $V_{\mathcal F^+}\geq 2$. Let
$0<\varepsilon<M^*/4$, we have
$$
      \mathcal N(\varepsilon,\mathcal F)\leq
      3\left(\frac{2eM^*}{\varepsilon}\log\frac{3eM^*}{\varepsilon}\right)^{V_\mathcal
      F^+}.
$$
\end{lemma}

\begin{lemma}\label{Lemma:covering number for summation}
Let $\mathcal F$ and $\mathcal G$ be two families of real functions.
If $\mathcal F\oplus\mathcal G$ denotes the set of functions
$\{f+g:f\in\mathcal F, g\in\mathcal G\}$, then for any
$\varepsilon,\nu>0$, we have
$$
      \mathcal N(\varepsilon+\nu,\mathcal F\oplus\mathcal G)
      \leq
      \mathcal N(\varepsilon,\mathcal F)\mathcal N(\nu,\mathcal G).
$$
\end{lemma}

\begin{IEEEproof}
 Let $\{f_1,\dots,f_N\}$ and $\{g_1,\dots,g_L\}$ be an
 $\varepsilon$-cover and a $\nu$-cover of $\mathcal F$ and $\mathcal
 G$, respectively. Then, for every $f\in\mathcal F$ and
 $g\in\mathcal G$, there exist $k\in\{1,\dots,N\}$ and
 $\ell\in\{1,\dots,L\}$ such that
$$
     \|f-f_k\|_\infty<\varepsilon,\qquad \|g-g_\ell\|_\infty<\nu.
$$
Due to the triangle inequality, we have
$$
     \|f+g-f_k-g_\ell\|_\infty\leq
     \|f-f_k\|_\infty+\|g-g_\ell\|_\infty\leq\varepsilon+\nu,
$$
which shows that $\{f_k+g_\ell:1\leq k\leq N,1\leq\ell\leq L\}$ is
an $(\varepsilon+\nu)$-cover of $\mathcal F\oplus\mathcal G$. The
definition of covering number then yields
$$
      \mathcal N(\varepsilon+\nu,\mathcal F\oplus\mathcal G)
      \leq
      \mathcal N(\varepsilon,\mathcal F)\mathcal N(\nu,\mathcal G).
$$
 This finishes the proof
of Lemma \ref{Lemma:covering number for summation}.
\end{IEEEproof}

\begin{lemma}\label{Lemma:covering number for product}
Let $\mathcal F$ and $\mathcal G$ be two families of real functions
 uniformly bounded by $M_1$ and $M_2$ respectively. If $\mathcal F\odot\mathcal G$ denotes the set of functions
$\{f\cdot g:f\in\mathcal F, g\in\mathcal G\}$, then for any
$\varepsilon,\nu>0$, we have
$$
      \mathcal N(\varepsilon+\nu,\mathcal F\odot\mathcal G)
      \leq
      \mathcal N(\varepsilon/M_2,\mathcal F)\mathcal N(\nu/M_1,\mathcal G).
$$
\end{lemma}

\begin{IEEEproof} Let $\{f_1,\dots,f_N\}$ and $\{g_1,\dots,g_L\}$ be an
 $\varepsilon/M_2$-cover and a $\nu/M_1$-cover of $\mathcal F$ and $\mathcal
 G$, respectively.   Then, for every $f\in\mathcal F$ and
 $g\in\mathcal G$, there exist $k\in\{1,\dots,N\}$ and
 $\ell\in\{1,\dots,L\}$ such that $\|f_k\|_\infty\leq M_1$,
 $\|g_\ell\|_\infty\leq M_2$, and
$$
     \|f-f_k\|_\infty<\varepsilon/M_2,\qquad \|g-g_\ell\|_\infty<\nu/M_1.
$$
It then follows from the triangle inequality that
\begin{eqnarray*}
    &&\|fg-f_kg_\ell\|_\infty
     \leq
    \|fg-fg_\ell\|_\infty+\|fg_\ell-f_kg_\ell\|_\infty\\
    &\leq&
    M_1\|g-g_\ell\|_\infty+\|g_\ell\|\|f-f_k\|_\infty
     \leq
    \nu+\varepsilon,
\end{eqnarray*}
which implies that $\{f_kg_\ell:1\leq k\leq N,1\leq \ell\leq L\}$ is
an $(\varepsilon+\nu)$-cover of $\mathcal F\odot\mathcal G$. This
together with the definition of covering number  finishes the proof
of Lemma \ref{Lemma:covering number for product}.
\end{IEEEproof}

By the help of  previous lemmas, we are in a position to prove
Proposition \ref{Proposition:covering number}.

\begin{IEEEproof}[Proof of Proposition \ref{Proposition:covering number}] According to
Lemma \ref{Lemma:covering number for summation}, we have
\begin{equation}\label{covering 1}
     \mathcal N(\varepsilon, \Phi_{n,2d})
     \leq
     \left(\max_{1\leq j\leq n^d}\mathcal N(\varepsilon/n^d,\mathcal G_{1,j})\right)^{n^d},
\end{equation}
where
\begin{eqnarray*}
      \mathcal G_{1,j}:=\big\{g_j&:&
   |c_j|\leq\mathcal C_n, |b_{  j}|\leq \mathcal B_n,|\alpha_{j,\ell}|,\\
   &&|\alpha'_{j,\ell}|\leq \Xi_n,
     \beta_{j,\ell} , \gamma_{j,\ell}\in\mathbb R\big\},
\end{eqnarray*}
and
\begin{eqnarray*}
       g_j(x)&:=&c_j\sigma\left(\sum_{\ell=1}^{d}\alpha_{j,\ell}\sigma_0
     \left(x^{(\ell)}+\beta_{j,\ell}\right)\right.\\
     &+&
     \left.\sum_{\ell=1}^{d}\alpha'_{j,\ell}\sigma_0
     \left(x^{(\ell)}+\gamma_{j,\ell}\right)+b_j\right).
\end{eqnarray*}
 Since $|c_{j}|\leq \mathcal C_n$ for all $1\leq j\leq n^d$ and $\|\sigma\|_\infty\leq 1$, we
obtain from Lemma \ref{Lemma:covering number for product} that for
arbitrary $1\leq j\leq n^d$, there holds
\begin{equation}\label{covering 2}
   \mathcal N(\varepsilon/n^d,\mathcal G_{1,j})
   \leq
   \mathcal N(\varepsilon/n^d,\{c_j:|c_j|\leq \mathcal C_n\})\mathcal
   N(\varepsilon/(\mathcal C_nn^d),\mathcal G_{2,j}),
\end{equation}
where
$$
      \mathcal G_{2,j}:=\left\{h_j:
     |b_{  j}|\leq \mathcal B_n,|\alpha_{j,\ell}|,|\alpha'_{j,\ell}|\leq \Xi_n,
    \beta_{j,\ell} , \gamma_{j,\ell}\in\mathbb R\right\},
$$
 and
\begin{eqnarray*}
       h_j(x)&:=& \sigma\left(\sum_{\ell=1}^{d}\alpha_{j,\ell}\sigma_0
     \left(x^{(\ell)}+\beta_{j,\ell}\right)\right.\\
     &+&
     \left.\sum_{\ell=1}^{d}\alpha'_{j,\ell}\sigma_0
     \left(x^{(\ell)}+\gamma_{j,\ell}\right)+b_j\right).
\end{eqnarray*}
 From the definition of the covering number, we can deduce
\begin{equation}\label{covering 3}
      \mathcal N(\varepsilon/n^d,\{c_j:|c_j|\leq \mathcal C_n\})
      \leq\frac{2\mathcal C_n}{\varepsilon/n^d}=\frac{2\mathcal C_nn^d}{\varepsilon}.
\end{equation}
Due to (\ref{lip for sigma}), we get
$$
      \|\sigma(f_1(\cdot))-\sigma(f_2(\cdot))\|_\infty
      \leq C_\sigma\|f_1-f_2\|_\infty,
$$
which implies
\begin{equation}\label{covering 4}
       \mathcal
   N(\varepsilon/(\mathcal C_nn^d),\mathcal G_{2,j})
   \leq
    \mathcal
   N(\varepsilon/(C_\sigma \mathcal C_nn^d),\mathcal G_{3,j}),
\end{equation}
where
$$
      \mathcal G_{3,j}:=\left\{p_j:
     |b_{  j}|\leq \mathcal B_n,|\alpha_{j,\ell}|,|\alpha'_{j,\ell}|\leq \Xi_n,
    \beta_{j,\ell} , \gamma_{j,\ell}\in\mathbb R \right\},
$$
 and
$$
       p_j:= \sum_{\ell=1}^{d}\alpha_{j,\ell}\sigma_0
     \left(x^{(\ell)}+\beta_{j,\ell}\right)+\sum_{\ell=1}^{d}\alpha'_{j,\ell}\sigma_0
     \left(x^{(\ell)}+\gamma_{j,\ell}\right)+b_j.
$$
Lemma \ref{Lemma:covering number for summation} then implies
\begin{eqnarray}\label{covering 5}
     && \mathcal
   N\left(\frac{\varepsilon}{C_\sigma \mathcal C_nn^d},\mathcal
   G_{3,j}\right)\nonumber\\
    &\leq&
   \mathcal N\left(\frac{\varepsilon}{(2d+1)C_\sigma \mathcal C_nn^d},\{b_j:|b_j|\leq \mathcal
   B_n\}\right) \nonumber\\
   &\times&
   \left[\max_{1\leq\ell\leq d}\mathcal N\left(\frac{\varepsilon}{(2d+1)C_\sigma \mathcal C_n
   n^d},\mathcal
   G_{4,j,\ell}\right)\right]^{d}\nonumber\\
   &\times&\left[\max_{1\leq\ell\leq d}\mathcal
N\left(\frac{\varepsilon}{(2d+1)C_\sigma \mathcal C_n
   n^d},\mathcal
   G'_{4,j,\ell}\right)\right]^{d},
\end{eqnarray}
where
$$
     \mathcal G_{4,j,\ell}:=\left\{
          \alpha_{j,\ell}\sigma_0
     \left(x^{(\ell)}+\beta_{j,\ell}\right):
   |\alpha_{j,\ell}|\leq \Xi_n,  \gamma_{j,\ell}\in\mathbb R\right\},
$$
and
$$
     \mathcal G'_{4,j,\ell}:=\left\{
          \alpha'_{j,\ell}\sigma_0
     \left(x^{(\ell)}+\gamma_{j,\ell}\right):
   |\alpha'_{j,\ell}|\leq \Xi_n,\beta_{j,\ell}\in\mathbb R\right\}.
$$
 From the definition of the covering number again, we can deduce
\begin{eqnarray}\label{covering 6}
       &&\mathcal N\left(\frac{\varepsilon}{(2d+1)C_\sigma \mathcal C_nn^d},\{b_j:|b_j|
       \leq \mathcal B_n\}\right)\nonumber\\
      &\leq& \frac{2\mathcal B_n(2d+1)C_\sigma \mathcal C_n
      n^{d}}{\varepsilon}.
\end{eqnarray}
Furthermore, it follows from Lemma \ref{Lemma:covering number for
product} and $|\alpha_{j,\ell}|,|\alpha'_{j,\ell}|\leq \Xi_n$,
$\|\sigma_0\|_\infty\leq 1$ that
\begin{eqnarray}\label{covering 7}
    &&\mathcal N\left(\frac{\varepsilon}{(2d+1)C_\sigma \mathcal C_nn^d},\mathcal
   G_{4,j,\ell}\right)\nonumber\\
   &\leq& \mathcal N\left(\frac{\varepsilon}{(2d+1)C_\sigma \mathcal C_nn^{d}},
   \{\alpha_{j,\ell}:|\alpha_{j,\ell}|\leq
   \Xi_n\}\right) \nonumber\\
   &\times&
   \mathcal N\left(\frac{\varepsilon}{(2d+1)\mathcal C_nC_\sigma n^{d}\Xi_n},
   \mathcal G_{5,j,\ell}\right),
\end{eqnarray}
and
\begin{eqnarray}\label{covering 7.1}
    &&\mathcal N\left(\frac{\varepsilon}{(2d+1)C_\sigma \mathcal C_nn^d},\mathcal
   G'_{4,j,\ell}\right)\nonumber\\
   &\leq& \mathcal N\left(\frac{\varepsilon}{(2d+1)C_\sigma \mathcal C_nn^{d}},
   \{\alpha_{j,\ell}:|\alpha_{j,\ell}|\leq
   \Xi_n\}\right)\nonumber\\
   &\times&
   \mathcal N\left(\frac{\varepsilon}{(2d+1)\mathcal C_nC_\sigma n^{d}\Xi_n},
   \mathcal G'_{5,j,\ell}\right),
\end{eqnarray}
 where
$$
     \mathcal G_{5,j,\ell}:=\left\{
          \sigma_0
     \left(x^{(\ell)}+\beta_{j,\ell}\right):
   \beta_{j,\ell}\in\mathbb R\right\},
$$
and
$$
     \mathcal G'_{5,j,\ell}:=\left\{
          \sigma_0
     \left(x^{(\ell)}+\gamma_{j,\ell}\right):
     \gamma_{j,\ell}\in\mathbb R\right\}.
$$
 Similarly, it is easy to see
\begin{eqnarray}\label{covering 8}
         &&\mathcal N\left(\frac{\varepsilon}{(2d+1)C_\sigma \mathcal C_nn^{d}\Xi_n},
         \{\alpha_{j,\ell}:|\alpha_{j,\ell}|\leq
   \Xi_n\}\right)\nonumber\\
   &\leq& \frac{2\Xi_n(2d+1)C_\sigma \mathcal C_nn^{d}}{\varepsilon}.
\end{eqnarray}
To bound the covering numbers of $\mathcal G_{5,j,\ell}$ and
$\mathcal G_{5,j,\ell}'$, we notice that $\sigma_0$ is a
non-decreasing function.  Then,   it follows from Lemma
\ref{Lemma:relation vc dimension} that $
        V_{\mathcal G_{5,j,\ell}^+}\leq V_{\mathcal G_{6,j,\ell}^+},
$ where
$$
           \mathcal G_{6,j,\ell}:=\left\{
            x^{(\ell)}+\beta_{j,\ell}:
    \beta_{j,\ell}\in\mathbb R\right\}.
$$
Noting  $\mathcal G_{6,j,\ell}$ is in a  one-dimensional linear
space, the definition of $\mathcal G_{6,j,\ell}^+$ implies
$$
      \mathcal G_{6,j,\ell}^+\subseteq\left\{\{(z,t)\in\mathbb R\times\mathbb
      R:\alpha t+g(z)\geq0\}:g\in\mathcal G_{6,j,\ell},\alpha\in\mathbb
      R\right\}
$$
and thus $\mathcal G_{6,j,\ell}^+$ is in a two-dimensional linear
space. Therefore, it follows from Lemma
  \ref{Lemma:esitmate vc dimenision} that $V_{\mathcal G_{6,j,\ell}^+}\leq
  2,$
which implies $
        V_{\mathcal G_{5,j,\ell}^+}\leq 2.
$ Therefore, it follows from Lemma \ref{lemma:inner covering number}
with $M^*=1$ that
\begin{eqnarray}\label{covering 9}
      &&\mathcal N\left(\frac{\varepsilon}{(2d+1)\mathcal C_nC_\sigma n^{d}\Xi_n},\mathcal
      G_{5,j,\ell}\right)\\
      &\leq&
      3\left(\frac{2e(2d+1)\mathcal C_nC_\sigma n^{d}\Xi_n}{\varepsilon}
      \log\frac{3e(2d+1)\mathcal C_nC_\sigma
      n^{d}\Xi_n}{\varepsilon}\right)^2. \nonumber
\end{eqnarray}
The same method also yields
\begin{eqnarray}\label{covering 9.1}
      &&\mathcal N\left(\frac{\varepsilon}{(2d+1)\mathcal C_nC_\sigma n^{d}\Xi_n},\mathcal
      G'_{5,j,\ell}\right)\\
      &\leq&
      3\left(\frac{2e(2d+1)\mathcal C_nC_\sigma n^{d}\Xi_n}{\varepsilon}
      \log\frac{3e(2d+1)\mathcal C_nC_\sigma n^{d}\Xi_n}{\varepsilon}\right)^2.\nonumber
\end{eqnarray}
Plugging (\ref{covering 9}) and (\ref{covering 8}) into
(\ref{covering 7}) and inserting (\ref{covering 9.1}) and
(\ref{covering 8}) into (\ref{covering 7.1}), we obtain
\begin{eqnarray*}
    &&\mathcal N\left(\frac{\varepsilon}{(2d+1)C_\sigma \mathcal C_nn^d},\mathcal
   G_{4,j,\ell}\right)\\
   &\leq&
   \frac{6\Xi_n(2d+1)C_\sigma \mathcal C_nn^{d}}{\varepsilon}\\
   &\times&
   \left(\frac{2e\mathcal C_nC_\sigma n^{d}\Xi_n}{\varepsilon}
      \log\frac{3e(2d+1)\mathcal C_nC_\sigma n^{d}\Xi_n}{\varepsilon}\right)^2\\
      &=&
      \frac{24e^2(2d+1)^3\mathcal C_n^3C^3_\sigma
      n^{3d}\Xi_n^3}{\varepsilon^3}\\
      &\times&\left(\log\frac{3e(2d+1)\mathcal C_nC_\sigma
      n^{d}\Xi_n}{\varepsilon}\right)^2
\end{eqnarray*}
and
\begin{eqnarray*}
    &&\mathcal N\left(\frac{\varepsilon}{(2d+1)C_\sigma \mathcal C_nn^d},\mathcal
   G'_{4,j,\ell}\right)\\
    &\leq&
      \frac{24e^2(2d+1)^3\mathcal C_n^3C^3_\sigma
      n^{3d}\Xi_n^3}{\varepsilon^3}\\
      &\times&
      \left(\log\frac{3e(2d+1)\mathcal C_nC_\sigma
      n^{d}\Xi_n}{\varepsilon}\right)^2.
\end{eqnarray*}
 Inserting the above two estimates and (\ref{covering 6}) into
(\ref{covering 5}), we then get
\begin{eqnarray*}
    && \mathcal
   N\left(\frac{\varepsilon}{C_\sigma \mathcal C_nn^d},\mathcal G_{3,j}\right)\leq
   \frac{2\mathcal B_n(2d+1)C_\sigma \mathcal C_n
      n^{d}}{\varepsilon}\\
      &\times&
   \left[\frac{24e^2(2d+1)^3\mathcal C_n^3C^3_\sigma
   n^{3d}\Xi_n^3}{\varepsilon^3}\right.\\
   &\times&\left.
      \left(\log\frac{3e(2d+1)\mathcal C_nC_\sigma
      n^{d}\Xi_n}{\varepsilon}\right)^2\right]^{ 2d }.
\end{eqnarray*}
This together with (\ref{covering 4}), (\ref{covering 3}) and
(\ref{covering 2}) yields
\begin{eqnarray*}
  &&\mathcal N\left(\frac{\varepsilon}{n^d},\mathcal G_{1,j}\right)
   \leq
      \left(\log\frac{3e (2d+1)\mathcal C_nC_\sigma
      n^{d}\Xi_n}{\varepsilon}\right)^{4d}\\
      &&\frac{4\mathcal B_n(24e^2)^{2d}(2d+1)^{6d+1}\mathcal C_n^{6d+2}\Xi_n^{6d}C_\sigma^{6d+1} n^{6d^2+2d}}{\varepsilon^{6d+2}}
\end{eqnarray*}
Plugging the above inequality into (\ref{covering 1}), we get
\begin{eqnarray*}
     &&\log\mathcal N(\varepsilon, \Phi_{n,2d})
      \leq
      4dn^d\log\log\frac{3e (2d+1)\mathcal C_nC_\sigma
      n^{d}\Xi_n }{\varepsilon}
      \\
     &+&
     n^d\log \frac{4\mathcal
      B_n(24e^2)^{2d}(2d+1)^{6d}
      \mathcal C_n^{6d+2}\Xi_n^{6d}C_\sigma^{6d+1} n^{6d^2+2d}}{\varepsilon^{6d+2}}.
 \end{eqnarray*}
This finishes the proof of Proposition \ref{Proposition:covering
number}.
\end{IEEEproof}

\section*{Appendix C: Deriving Learning Rates}
In this appendix, we aim at proving results in Section \ref{Sec.Main
result}. Our main idea is motivated by the classical error
decomposition strategy proposed \cite{Wu2006} that divides the
generalization error into the approximation error and sample error.
The approximation error can be estimated by using Propositions
\ref{Proposition:localization} and \ref{Proposition:sparse
approximation}, while the sample error is estimated by   using
Proposition \ref{Proposition:covering number} and some concentration
inequality in statistics.

\subsection{Error decomposition}

Define
 \begin{equation}\label{first order NN}
      N_{n,2d,L}(x)
      =
      \sum_{{\bf j}\in\mathbb N_n^d}f_\rho(\xi_{\bf j})N^*_{n,{\bf
      j},L}(x)
\end{equation}
with $L$ being defined by (\ref{definition K for sigmoidal 222}).
Since $|y_i|\leq M$ almost surely,  it follows from Assumption
\ref{Ass:space} that $N_{n,2d,L}\in\Phi_{n,2d}$. The follow lemma
presents the error decomposition for our analysis.

\begin{lemma}\label{Lemma:error decomposition}
Let $f_{D,n}$ and $N_{n,2d,L}$ be defined by (\ref{ERM}) and
(\ref{first order NN}), respectively. Then, we have
\begin{eqnarray*}
       &&\mathcal E(\pi_Mf_{D,n})-\mathcal E(f_\rho)
       \leq \mathcal E(N_{n,2d,L})-\mathcal E(f_\rho)
       \\
       &+&\mathcal E(\pi_Mf_{D,n}) -
       \mathcal E_D(\pi_Mf_{D,n})\\
       &+&\mathcal
       E_D(N_{n,2d,L})-\mathcal E(N_{n,2d,L}),
\end{eqnarray*}
where $
      \mathcal E_D(f)=\frac1m\sum_{i=1}^m
      (f(x_i)-y_i)^2.
$
\end{lemma}

\begin{IEEEproof}
 It is obvious that
\begin{eqnarray*}
       &&\mathcal E(\pi_Mf_{D,n})-\mathcal E(f_\rho)
        \leq
        \mathcal E(N_{n,2d,L})-\mathcal E(f_\rho)\\
       &+&\mathcal E(\pi_Mf_{D,n})
        - \mathcal E_D(\pi_Mf_{D,n}) +
       \mathcal
       E_D(N_{n,2d,L})\\
       &-&\mathcal E(N_{n,2d,L})
         +  \mathcal E_D(\pi_Mf_{D,n})-\mathcal
       E_D(N_{n,2d,L}).
\end{eqnarray*}
Due to the definition of $\pi_M$, it follows from (\ref{ERM}) and
$N_{n,2d,L}\in\Phi_{n,2d}$ that
$$
        \mathcal E_D(\pi_Mf_{D,n})-\mathcal
       E_D(N_{n,2d,L})\leq \mathcal E_D(f_{D,n})-\mathcal
       E_D(N_{n,2d,L})\leq0.
$$
  This finishes the proof of
Lemma \ref{Lemma:error decomposition}.
\end{IEEEproof}

Setting $\mathcal D_n:=\mathcal E(N_{n,2d,L})-\mathcal E(f_\rho)$,
$\mathcal S_1:=\mathcal S_{1,n,D}:=\mathcal
       E_D(N_{n,2d,L})-\mathcal E(N_{n,2d,L})$
and $\mathcal S_2:=\mathcal S_{2,n,D}:=\mathcal
E(\pi_Mf_{D,n})-\mathcal E_D(\pi_Mf_{D,n})$,   we get from Lemma
\ref{Lemma:error decomposition} that
\begin{equation}\label{error decomposition 111}
       \mathcal E(\pi_Mf_{D,n})-\mathcal E(f_\rho)
       \leq
       \mathcal D_n+\mathcal S_1+\mathcal S_2.
\end{equation}

\subsection{Approximation error estimate}
%

The main tool to present the approximation error estimate is
Proposition \ref{Proposition:sparse approximation}. Indeed, we can
deduce the following tight bounds for $\mathcal D_n$.

\begin{proposition}\label{Proposition:first order approximation}
 Under
Assumptions \ref{Ass:sigmoidal}, \ref{Ass:regression},
  \ref{Ass:space}, there holds
\begin{equation}\label{App pro 1}
      \mathcal D_n\leq (2^{r}c_0^2+M^2)n^{-2r}.
\end{equation}
 Under Assumptions \ref{Ass:sigmoidal}, \ref{Ass:regression1}, \ref{Ass:margin},
\ref{Ass:space},  there holds
\begin{equation}\label{App pro 2}
      \mathcal D_n\leq (c_12^{r+d}c_0^2+(1+c_1)M^2)n^{-2r}\frac{s}{N^d}.
\end{equation}
\end{proposition}

\begin{IEEEproof}
 Due to (\ref{equality}) and
$\|\cdot\|_\rho\leq\|\cdot\|_{L^\infty(\mathbb I)^d}$, we have
$$
     \mathcal D_n=\|f_\rho-N_{n,2d,L}\|_\rho^2\leq \|f_\rho-N_{n,2d,L}\|_{L^\infty(\mathbb I)^d}^2.
$$
Then, it follows from $\|f_\rho\|_{L^\infty(\mathbb I)^d}\leq M$,
(\ref{sparse approximation1}) with $\eta=\{\xi_{\bf j}\}_{{\bf
j}\in\mathbb N_n^d}$, $K_\varepsilon=L$ and $\varepsilon=n^{-r-d}$
that
$$
     \mathcal D_n\leq (2^{r}c_0^2+M^2)n^{-2r},
$$
which proves (\ref{App pro 1}).

Now, we turn to bound (\ref{App pro 2}). It is easy to check that
\begin{eqnarray}\label{app decom 111}
   &&\mathcal D_n=\int_{\mathcal
   X}|f_\rho(x)-N_{n,2d,L}(x)|^2d\rho_X \nonumber \\
   &\leq&
   \sum_{{\bf k}\in \Lambda_s}\sum_{{\bf j}\in \overline{\Lambda_{\bf
   k}}}\int_{A_{n,{\bf j}}}|f_\rho(x)-N_{n,2d,L}(x)|^2d\rho_X \nonumber \\
   &+&
   \sum_{{\bf k}\in \Lambda_s}\sum_{{\bf j}\notin \overline{\Lambda_{\bf
   k}}}\int_{A_{n,{\bf j}}}|f_\rho(x)-N_{n,2d,L}(x)|^2d\rho_X  \nonumber\\
   &=:&
   \mathcal J_1+\mathcal J_2.
\end{eqnarray}
From (\ref{Card over lam}) and (\ref{sparse approximation1})
Assumption \ref{Ass:margin} and Assumption \ref{Ass:space}, we get
\begin{eqnarray*}
    \mathcal J_1
    & \leq&
    (2^{r}c_0^2+M^2)n^{-2r}\sum_{{\bf k}\in \Lambda_s}\sum_{{\bf j}\notin \overline{\Lambda_{\bf
   k}}}\int_{A_{n,{\bf j}}}d\rho_X\\
   &\leq&
   c_1(2^{r+d}c_0^2+M^2)n^{-2r}\frac{s}{N^d}.
\end{eqnarray*}
  Since
$n\geq 4N$, we get from Assumption \ref{Ass:margin}, Assumption
 (\ref{sparse approximation2}) with $\varepsilon=$ that
$$
    \mathcal J_2\leq  M^2n^{2d}\varepsilon^2\leq M^2n^{-2r}\frac{s}{N^d}.
$$
Plugging the above two estimates into (\ref{app decom 111}), we get
$$
      \mathcal D_n\leq
     (c_12^{r+d}c_0^2+(1+c_1)M^2)n^{-2r}\frac{s}{N^d}.
$$
This completes the proof of Proposition \ref{Proposition:first order
approximation}.
\end{IEEEproof}

\subsection{Sample error estimate}

To bound $\mathcal S_1$, we need the following two Lemmas. The first
is the  Bernstein inequality, which was proved in \cite{Shi2011}.

\begin{lemma}\label{Lemma:Bernstein}
 Let $\xi$ be a random variable on a probability space
$\mathcal Z$ with variance $\sigma^2$ satisfying $|\xi-\mathbf
E\xi|\leq M_\xi$ for some constant $M_\xi$. Then for any
$0<\delta<1$, with confidence $1-\delta$, we have
$$
             \frac1m\sum_{i=1}^m\xi(z_i)-\mathbf
             E\xi\leq\frac{2M_\xi\log\frac1\delta}{3m}+\sqrt{\frac{2\sigma^2\log\frac1\delta}{m}}.
$$
\end{lemma}

The second lemma presents a bound for the summation of $N^*_{n,{\bf
j},L}$.

\begin{lemma}\label{Lemma: bound for sum}
   Let $N^*_{n,{\bf j},L}$ be defined by (\ref{NN for localization}) with $L$ satisfying
(\ref{definition K for sigmoidal 222}).
   Under  Assumption \ref{Ass:sigmoidal}, there holds
$$
       \sum_{{\bf j}\in \mathbb N_n^d}\left|N^*_{n,{\bf j},L}(x)\right|\leq 2^d+1, \qquad\forall
       x\in[0,1]^d.
$$
\end{lemma}

\begin{IEEEproof}
   Due to the definition of $ A_{n,{\bf j}}$, we have $
      [0,1]^d=\bigcup_{{\bf j}\in\mathbb N_n^d} A_{n,{\bf j}}.$
      Furthermore, it is easy to see that for arbitrary $x\in
      [0,1]^d$, there are at most $2^d$ ${\bf j}$'s denoted by ${\bf j}_1,\dots,{\bf j}_{2^d}$
       such that $x\in  A_{n,{\bf
      j}_k}, k=1,\dots, 2^d$.
Then it follows from Proposition \ref{Proposition:localization} that
\begin{eqnarray*}
       &&\sum_{{\bf j}\in \mathbb N_n^d}\left|N^*_{n,{\bf
       j},L}(x)\right|=\sum_{k=1}^{2^d}\left|N^*_{n,{\bf
       j}_k,L}(x)\right|\\
       &+&
       \sum_{{\bf j}\neq{\bf j}_1,\dots,{\bf j}_{2^d}}\left|N^*_{n,{\bf
       j},L}(x)\right|
       \leq
       2^d+n^dn^{-s-d}\leq 2^d+1.
\end{eqnarray*}
This finishes the proof of Lemmas \ref{Lemma: bound for sum}.
\end{IEEEproof}

By the help of the above lemma, we obtain the following Proposition
\ref{Proposition:Bound S1}.

\begin{proposition}\label{Proposition:Bound S1}
 For any $0<\delta<1$, with confidence
$1-\frac\delta2$,
\begin{eqnarray*}
       \mathcal
       S_1
        \leq
       \frac{7M^2(2^d+4)^2\log\frac2\delta}{3m}+\frac12\mathcal D_n.
\end{eqnarray*}
\end{proposition}

\begin{IEEEproof}
 Let the random variable $\xi$ on $\mathcal Z$ be defined by
$$
        \xi({  z})=(y-N_{n,2d,L}(x))^2-(y-f_\rho(x))^2 \quad {  z}=(x,y)\in
             \mathcal Z.
$$
Since $|f_\rho(x)|\leq M$ almost everywhere, it follows from Lemma
\ref{Lemma: bound for sum} that
\begin{eqnarray*}
          |\xi({  z})|
          &=&
          |(f_\rho(x)-N_{n,2d,L}(x))(2y-N_{n,2d}(x)-f_\rho(x))|\\
          &\leq&
          M^2(2^d+2)(2^d+4)
          \leq
           M_\xi:=M^2(2^d+4)^2
\end{eqnarray*}
and almost surely
$$
            |\xi-\mathbf E\xi|\leq 2M_\xi.
$$
Moreover, we have
\begin{eqnarray*}
            &&\mathbf E(\xi^2)\\
             &=&
            \int_Z(N_{n,2d,L}(x)+f_\rho(x)-2y)^2(N_{n,2d,L}-f_\rho(x))^2d\rho\\
             &\le&
             M_\xi\|f_\rho-N_{n,2d,L}\|^2_\rho,
\end{eqnarray*}
which implies that the variance $\sigma^2$ of $\xi$ can be bounded
as $\sigma^2\leq \mathbf E(\xi^2)\leq M_\xi\mathcal D_n.$ Now
applying Lemma \ref{Lemma:Bernstein}, with confidence
$1-\frac\delta2$, we have
\begin{eqnarray*}
       \mathcal
       S_1
        &=&
       \frac1m\sum_{i=1}^m\xi(z_i)-\mathbf E\xi
        \leq
       \frac{4M_\xi\log\frac2\delta}{3m}+\sqrt{\frac{2M_\xi\mathcal
       D(n)\log\frac{2}{\delta}}{m}}\\
        &\leq&
       \frac{7M^2(2^d+4)^2\log\frac2\delta}{3m}+\frac12\mathcal D_n.
\end{eqnarray*}
This finishes the proof of Proposition \ref{Proposition:Bound S1}.
\end{IEEEproof}

To bound $\mathcal S_2$, we need  the following ratio probability
inequality which is a standard result in learning theory
\cite{Wu2006}.

\begin{lemma}\label{Lemma: CONCENTRATION INEQUALITY}
 Let $\mathcal G$ be a set of functions on $\mathcal Z$ such that, for
some $c\geq 0$, $|g-\mathbf E(g)|\leq B_0$ almost everywhere and
$\mathbf E(g^2)\leq c\mathbf E(g)$ for each $g\in\mathcal G$. Then,
for every $\varepsilon>0$,
\begin{eqnarray*}
             &&\mathbf P\left\{\sup_{f\in\mathcal G}
             \frac{\mathbf E(g)-\frac1m\sum_{i=1}^mg(z_i)}{\sqrt{ \mathbf E(g)+\varepsilon}}
             \geq\sqrt{\varepsilon}\right\}\\
             &\leq&
             \mathcal N(\varepsilon,\mathcal G)
             \exp\left\{-\frac{m\varepsilon}{2c+\frac{2B_0}3}\right\}.
\end{eqnarray*}
\end{lemma}

Using the above lemma and Proposition \ref{Proposition:covering
number}, we can deduce the following   estimate for $\mathcal S_2$.

\begin{proposition}\label{Proposition:Bound S2}
Let $0<\delta<1$. With confidence at least $1-\frac{\delta}2$, there
holds
\begin{eqnarray*}
     && \mathcal S_2\leq \frac12[\mathcal E(\pi_Mf_{D,n})-\mathcal E(f_\rho)]
      +m^\frac{-2s}{2s+d}428(6d+2)M^2  \\
      &\times&\log\left[192e^2(2d+1)MC_\sigma \mathcal
       B_n
      \mathcal C_n\Xi_nm\right]\log\frac2\delta.
\end{eqnarray*}
\end{proposition}

\begin{IEEEproof}
 Set
$$
             \mathcal F_n:=\{(\pi_Mf(x)-y)^2-(f_\rho(x)-y)^2:f\in
             \Phi_{n,2d}\}.
$$
Then for $g\in\mathcal F_n,$  there exists $f\in \Phi_{n,2d}$ such
that $g(z)=(\pi_Mf(x)-y)^2-(f_\rho(x)-y)^2$. Therefore,
$$
            \mathbf  E(g)=\mathcal E(\pi_Mf)-\mathcal E(f_\rho)\geq0,
$$
and
$$
              \frac1m\sum_{i=1}^mg(z_i)=\mathcal E_D(\pi_Mf)-\mathcal E_D(f_\rho).
$$
Since $|\pi_Mf|\leq M$ and $|f_\rho(x)|\leq M$ almost everywhere, we
find that
$$
             |g({
             z})|=|(\pi_Mf(x)-f_\rho(x))((\pi_Mf(x)-y)+(f_\rho(x)-y))|\leq8M^2,
$$
which together with (\ref{equality})  follows   $|g({z})-\mathbf
E(g)|\leq16M^2$ almost everywhere and
$$
             \mathbf E(g^2)
             \leq
             16M^2\|\pi_Mf-f_\rho\|^2_{L^2_{\rho}}=16M^2\mathbf E(g).
$$
Now we apply Lemma \ref{Lemma: CONCENTRATION INEQUALITY} with
$B_0=c=16M^2$ to the set of functions $\mathcal F_n$ and obtain that
\begin{eqnarray}\label{sample 1}
             \sup_{f\in \Phi_{n,2d}}\frac{\{\mathcal E(\pi_Mf)-\mathcal E(f_\rho)\}
             -\{\mathcal E_{D}(\pi_Mf)-\mathcal E_{D}(f_\rho)\}}{
             \sqrt{\{\mathcal E(\pi_Mf)-\mathcal
             E(f_\rho)\}+\varepsilon}}
             \leq
             \sqrt{\varepsilon}
\end{eqnarray}
with confidence at least
$$
             1-\mathcal N(\varepsilon,\mathcal F_n)\mbox{exp}\left\{-\frac{3m\varepsilon}{128M^2}\right\}.
$$

Observe that for $g_1,g_2\in\mathcal F_n$ there exist $f_1,f_2\in
\Phi_{n,2d}$ such that
$$
             g_j(z)=(\pi_Mf_j(x)-y)^2-(f_\rho(x)-y)^2,\ j=1,2.
$$
Then
\begin{eqnarray*}
             &&|g_1({ z})-g_2({ z})|
              =
             |(\pi_Mf_1(x)-y)^2-(\pi_Mf_2(x)-y)^2|\\
             &\leq&
             4M\|\pi_Mf_1-\pi_Mf_2\|_\infty\leq4M\|f_1-f_2\|_\infty.
\end{eqnarray*}
 We see that for any $\varepsilon>0$, an
$\left(\frac{\varepsilon}{4M}\right)$-covering of $\Phi_{n,2d}$
provides an $\varepsilon$-covering of $\mathcal F_n$. Therefore
$$
             \mathcal N(\varepsilon,\mathcal F_n)\leq
             \mathcal N\left(\frac{\varepsilon}{4M},\Phi_{n,2d}\right).
$$
Then the confidence is
\begin{eqnarray*}
             &&1-\mathcal N(\varepsilon,\mathcal
             F_n)\exp\left\{-\frac{3m\varepsilon}{128M^2}\right\}\\
              &\geq&
             1-\mathcal N\left(\frac{\varepsilon}{4M},\Phi_{n,2d}
             \right)\exp\left\{-\frac{3m\varepsilon}{128M^2}\right\}.
\end{eqnarray*}
According to Proposition \ref{Proposition:covering number}, we have
\begin{eqnarray*}
     &&\log\mathcal N(\varepsilon/4M, \Phi_{n,2d})\\
      &\leq&
      4dn^d\log\log\frac{12Me (2d+1)C_\sigma \mathcal C_n
      n^{d}\Xi_n }{\varepsilon}\\
      &+&(6d+2)n^d
      \log \big[\frac{M(4\mathcal B_n)^{\frac1{6d+2}}(24e^2)^{\frac{2d}{6d+2}}
      (2d+1)^{\frac{6d}{6d+2}}}{\varepsilon}\\
      &\times&
      \mathcal C_n\Xi_n^{\frac{6d}{6d+2}}C_\sigma^{\frac{6d+1}{6d+2}}
      n^{d}\big].
 \end{eqnarray*}
Thus it follows from the above estimate and (\ref{sample 1}) that,
with confidence  at least
\begin{eqnarray}\label{sample 2}
         &&1-\exp\big\{4dn^d\log\log\frac{12Me (2d+1)C_\sigma \mathcal C_n
      n^{d}\Xi_n }{\varepsilon}\nonumber\\
      &-&\frac{3m\varepsilon}{128M^2}
       +
      (6d+2)n^d\log \big[\frac{M(4\mathcal
      B_n)^{\frac1{6d+2}}(24e^2)^{\frac{2d}{6d+2}}}{\varepsilon}\nonumber\\
      &\times&
      (2d+1)^{\frac{6d}{6d+2}}
      \mathcal C_n\Xi_n^{\frac{6d}{6d+2}}C_\sigma^{\frac{6d+1}{6d+2}}
      n^{d}\big]\big\}
\end{eqnarray}
there holds
\begin{eqnarray*}
             &&\frac{\{\mathcal E(\pi_Mf_{D,n})-\mathcal E(f_\rho)\}
             -\{\mathcal E_{D}(\pi_Mf_{D,n})-\mathcal E_{D}(f_\rho)\}}{
             \sqrt{\{\mathcal E(\pi_Mf_{D,n})-\mathcal
             E(f_\rho)\}+\varepsilon}}\\
             &\leq&
             \sup_{f\in \Phi_{n,2d}}\frac{\{\mathcal E(\pi_Mf)-\mathcal E(f_\rho)\}
             -\{\mathcal E_{D}(\pi_Mf)-\mathcal E_{D}(f_\rho)\}}{
             \sqrt{\{\mathcal E(\pi_Mf)-\mathcal
             E(f_\rho)\}+\varepsilon}}\\
             &\leq&
             \sqrt{\varepsilon}.
\end{eqnarray*}
That is,
\begin{equation}\label{sample 3}
      \mathcal S_2\leq \frac12[\mathcal E(\pi_Mf_{D,n})-\mathcal E(f_\rho)]
      +\varepsilon.
\end{equation}
Define
\begin{eqnarray*}
   &&h(\eta):=4dn^d\log\log\frac{12Me (2d+1)C_\sigma \mathcal C_n
      n^{d}\Xi_n }{\varepsilon}\\
      &-&
      \frac{3m\varepsilon}{128M^2}
       +
      (6d+2)n^d\log\big[ \frac{M(4\mathcal
      B_n)^{\frac1{6d+2}}(24e^2)^{\frac{2d}{6d+2}}}{\varepsilon}\\
      &\times&
      (2d+1)^{\frac{6d}{6d+2}}
      \mathcal C_n\Xi_n^{\frac{6d}{6d+2}}C_\sigma^{\frac{6d+1}{6d+2}}
      n^{d}\big].
\end{eqnarray*}
Choose $\eta^*$ to be the positive solution to the equation
$$
              h(\eta)=\log\frac\delta2.
$$
The function
 $h:\mathbb R_+\rightarrow\mathbb R$
 is strictly
decreasing. Hence $\eta^*\leq\eta $ if $h(\eta )\leq
h(\eta^*)=\log\frac\delta2.$ Let $n=\left\lfloor
m^\frac{1}{2s+d}\right\rfloor$. For arbitrary  $\eta\geq
m^{-2s/(2s+d)}$, we have
\begin{eqnarray*}
              && h(\eta)
                \leq
                4dm^\frac{d}{2s+d}\log\log\left[12Me (2d+1)C_\sigma \mathcal C_n
      m\Xi_n
      \right]\\
      &-&
      \frac{3m\eta}{128M^2}
       +
               m^\frac{d}{2s+d}(6d+2)\log\\
               &&
                \left[M(4\mathcal
       B_n)^{\frac1{6d+2}}(24e^2)^{\frac{2d}{6d+2}}
      (2d+1)^{\frac{6d}{6d+2}}
      \mathcal C_n\Xi_n^{\frac{6d}{6d+2}}C_\sigma^{\frac{6d+1}{6d+2}} m\right].
\end{eqnarray*}
Take $\eta_1$ to be a positive number satisfying
\begin{eqnarray*}
               &&\log\frac\delta2=4dm^\frac{d}{2s+d}\log\log\left[12Me (2d+1)C_\sigma \mathcal C_n
      m\Xi_n
      \right]\\
      &-&\frac{3m\eta}{128M^2}
       +
               m^\frac{d}{2s+d}(6d+2) \log\big[M(4\mathcal
       B_n)^{\frac1{6d+2}}\\
               &&
      (24e^2)^{\frac{2d}{6d+2}}(2d+1)^{\frac{6d}{6d+2}}
      \mathcal C_n\Xi_n^{\frac{6d}{6d+2}}C_\sigma^{\frac{6d+1}{6d+2}} m\big].
\end{eqnarray*}
Then we have $h(\eta_1)\leq h(\eta^*)=\log\frac\delta2$, provided
$\eta_1\geq m^{-2s/(2r+s)}$. Direct computation yields
\begin{eqnarray*}
              &&\eta_1
               =
              \frac{512dM^2}{3}m^\frac{-2s}{2r+s}\log\log\left[12Me
              (2d+1)C_\sigma
      \mathcal C_n
      m\Xi_n
      \right]\\
      &+&
      \frac{128M^2}{3m}\log\frac{2}{\delta}
       +
              \frac{128M^2}{3}m^\frac{-2s}{2r+s}(6d+2) \log\big[M\\
              &&
              (4\mathcal
       B_n)^{\frac1{6d+2}}(24e^2)^{\frac{2d}{6d+2}}
      (2d+1)^{\frac{6d}{6d+2}}
       \mathcal C_n\Xi_n^{\frac{6d}{6d+2}}C_\sigma^{\frac{6d+1}{6d+2}}
      m\big].
\end{eqnarray*}
It is obvious that  $ \eta_1\geq m^{-2s/(2s+d)}.$ Then we obtain
\begin{eqnarray*}
       &&\eta^*\leq \eta_1  \leq
       43M^2m^{-1} \log\frac2\delta
        +
       m^\frac{-2s}{2s+d}214(6d+2)M^2\\
       &\times&
         \log\left[192e^2(2d+1)MC_\sigma  \mathcal
       B_n
      \mathcal C_n\Xi_nm\right].
\end{eqnarray*}
Hence, it follows from (\ref{sample 2}) and (\ref{sample 3}) that
with confidence at least $1-\frac{\delta}2$, there holds
\begin{eqnarray*}
      &&\mathcal S_2\leq \frac12[\mathcal E(\pi_Mf_{D,n})-\mathcal E(f_\rho)]
      +m^\frac{-2s}{2s+d}428(6d+2)M^2\\
      &\times&
        \log\left[192e^2(2d+1)MC_\sigma  \mathcal
       B_n
      \mathcal C_n\Xi_nm\right]\log\frac2\delta.
\end{eqnarray*}
This finishes the proof of Proposition \ref{Proposition:Bound S2}.
\end{IEEEproof}

\subsection{Learning rate analysis}
In this part, we prove results in Section \ref{Sec.Main result} by
using the error decomposition, approximation error estimate and
sample error estimate presented in the previous three subsections.

\begin{IEEEproof}[Proof of Theorem \ref{Theorem: ERM}] Due to (\ref{error
decomposition 111}) and Proposition \ref{Proposition:Bound S2},
there exists a subset ${\mathcal Z}_{\delta,1}^m$ of ${\mathcal
Z}^m$ with measure at least $1- \delta/2$ such that for every $D_m
\in {\mathcal Z}_{1, \delta}^{m}$, there holds
\begin{eqnarray}\label{error decomposition 333}
         &&\mathcal E(\pi_Mf_{D,n})-\mathcal E(f_\rho)
         \leq 2\mathcal D_n+2\mathcal S_1
          +
         856(6d+2)M^2\nonumber\\
         &\times&  \log\left[192e^2(2d+1)MC_\sigma  \mathcal
       B_n
      \mathcal C_n\Xi_nm\right]m^\frac{-2s}{2s+d}\log\frac2\delta.
\end{eqnarray}
Furthermore, it follows from Proposition \ref{Proposition:Bound S1}
that there exists a subset ${\mathcal Z}_{\delta,2}^m$ of ${\mathcal
Z}^m$ with measure at least $1- \delta/2$ such that for every $D_m
\in {\mathcal Z}_{2, \delta}^{m}$, there holds
\begin{equation}\label{abc}
    2\mathcal S_1\leq \frac{14M^2(2^d+4)^2\log\frac2\delta}{3m}+\mathcal
    D(n).
\end{equation}
Plugging the above estimate and (\ref{App pro 1}) into (\ref{error
decomposition 333}), and noting $n=\lfloor m^{1/(2s+d)}\rfloor$, we
have for every $D_m\in{\mathcal Z}_{\delta,1}^m\cap {\mathcal
Z}_{\delta,2}^m$, there holds
\begin{eqnarray*}
         &&\mathcal E(\pi_Mf_{D,n})-\mathcal E(f_\rho)
         \leq 2(2^{r}c_0^2+M^2)n^{-2r}m^{-\frac{2r}{2r+d}}\\
         &+&
         \frac{14M^2(2^d+4)^2\log\frac2\delta}{3m}
          +
         856(6d+2)M^2m^\frac{-2r}{2r+d}\log\frac2\delta\nonumber\\
         &\times&
           \log\left[192e^2(2d+1)MC_\sigma  \mathcal
       B_n
      \mathcal C_n\Xi_nm\right].
\end{eqnarray*}
Hence, with confidence at least $1-\delta$, there holds
$$
      \mathcal E(\pi_Mf_{D,n})-\mathcal E(f_\rho)
      \leq
      C\log\left[ \mathcal
       B_n
      \mathcal C_n\Xi_nm\right]m^\frac{-2s}{2s+d}\log\frac2\delta,
$$
where
\begin{eqnarray*}
     C&:=& 2(2^{r}c_0^2+M^2)+\frac{14M^2(2^d+4)^2}{3}\\
     &+& 856(12d+4)M^2\log(192e^2C_\sigma ).
\end{eqnarray*}
This finishes the proof of Theorem \ref{Theorem: ERM}.
\end{IEEEproof}

\begin{IEEEproof}[Proof of Corollary \ref{Corollary: optimal}] From the
confidence-based error bound (\ref{theorem1}), we obtain
 that the
nonnegative random variable $\xi =\mathcal E(\pi_Mf_{D,n})-\mathcal
E(f_\rho)$ satisfies
$$
         \mathbf P\left[\xi > t\right]
         \leq  2\exp\left\{-C^{-1}tm^{\frac{2s}{2s+d}}\log^{-1}(\mathcal
       B_n
      \mathcal C_n\Xi_n m)
          \right\}
$$
for any $t\geq C\log2m^\frac{-2s}{2s+d}\log\left[  \mathcal
       B_n
      \mathcal C_n\Xi_n m\right].$
Applying this bound to the formula
$$
      \mathbf E[\xi]=\int_0^\infty{\mathbf P}[\xi>t]dt
$$
for nonnegative random variables, we obtain
\begin{eqnarray*}
        &&\mathbf E
        \left[\mathcal E(\pi_Mf_{D,n})-\mathcal
        E(f_\rho)\right]
         \leq  C m^\frac{-2s}{2s+d}\log\left[  \mathcal
       B_n
      \mathcal C_n\Xi_n m\right]\log2 \\
        && +  2\int_0^\infty\exp\left\{-C^{-1}tm^{\frac{2s}{2s+d}}\log^{-1}(\mathcal
       B_n
      \mathcal C_n\Xi_n m)
          \right\}d t.
\end{eqnarray*}
By a change of variable, we see that the above integration equals
\begin{eqnarray*}
     && C m^\frac{-2s}{2s+d}\log(\mathcal
       B_n
      \mathcal C_n\Xi_n\Lambda_nm) \int_0^\infty  \exp\left\{-u\right\} du\\
      &=&C m^\frac{-2s}{2s+d}\log(\mathcal
       B_n
      \mathcal C_n\Xi_n m).
\end{eqnarray*}
Hence
$$
    \mathbf E
        \left[\mathcal E(\pi_Mf_{D,n})-\mathcal
        E(f_\rho)\right]
    \leq \left(2 + \log2\right)  C m^\frac{-2s}{2s+d}\log(\mathcal
       B_n
      \mathcal C_n\Xi_n m).
$$
This together with (\ref{baseline}) completes the  proof of
Corollary \ref{Corollary: optimal}.
\end{IEEEproof}

\begin{IEEEproof}[Proof of Theorem \ref{Theorem: ERM sparse}]
Plugging (\ref{error decomposition 333}), (\ref{abc})  and (\ref{App
pro 2}) into (\ref{error decomposition 333}), and noting
$n=\left\lfloor
\left(\frac{ms}{N^d}\right)^{\frac{d}{2r+d}}\right\rfloor$, we have
for every $D_m\in{\mathcal Z}_{\delta,1}^m\cap {\mathcal
Z}_{\delta,2}^m$, there holds
\begin{eqnarray*}
         &&\mathcal E(\pi_Mf_{D,n})-\mathcal E(f_\rho)\\
         &\leq&
          2(2^{r}c_0^2+2M^2)m^{-\frac{2r}{2r+d}}\left(\frac{s}{N^d}\right)^{\frac{d}{2r+d}}\\
         &+&
         \frac{14M^2(2^d+4)^2\log\frac2\delta}{3m}\\
          &+&
         856(6d+2)M^2m^\frac{-2r}{2r+d}\left(\frac{s}{N^d}\right)^{\frac{d}{2r+d}}\log\frac2\delta\nonumber\\
         &\times&
           \log\left[192e^2(2d+1)MC_\sigma  \mathcal
       B_n
      \mathcal C_n\Xi_nm\right].
\end{eqnarray*}
Hence, with confidence at least $1-\delta$, there holds
\begin{eqnarray*}
      &&\mathcal E(\pi_Mf_{D,n})-\mathcal E(f_\rho)\\
      &\leq&
      C'\log\left[ \mathcal
       B_n
      \mathcal C_n\Xi_nm\right]m^\frac{-2s}{2s+d}\left(\frac{s}{N^d}\right)^{\frac{d}{2r+d}}\log\frac2\delta,
\end{eqnarray*}
where
\begin{eqnarray*}
     C'&:=& 2(c_12^{r}c_0^2+(1+c_1)M^2)+\frac{14M^2(2^d+4)^2}{3}\\
     &+& 856(12d+4)M^2\log(192e^2C_\sigma ).
\end{eqnarray*}
This finishes the proof of Theorem \ref{Theorem: ERM sparse}.
\end{IEEEproof}

\begin{IEEEproof}[Proof of Corollary \ref{Corollary: optimal sparse}] The  bound can be deduced from the
confidence-based error bound in Theorem \ref{Theorem: ERM sparse} by
the same method as that in the proof of Corollary \ref{Corollary:
optimal}.
\end{IEEEproof}

  \section*{Acknowledgement}
The research was supported by the National Natural Science
Foundation of China (Grant Nos. 61502342, 11771012). The author
would like to thank  two anonymous referees for their constructive
suggestions. The author also would like to thank Professor Jinshan
Zeng for his helpful suggestions in revising the paper.

\end{document}